\documentclass{elsarticle}

\usepackage{pgfplots}
\usepackage{tikz}
\usepackage{subcaption}
\usepackage{algorithm}
\usepackage[]{algpseudocode}

\usepackage{amsthm,amsfonts,amsmath,amssymb}

\makeatletter
\renewcommand{\ALG@name}{Pseudocode}
\makeatother

\newtheorem{thm}{Theorem}

\newtheorem{lem}[thm]{Lemma}
\theoremstyle{definition}
\newtheorem{defn}[thm]{Definition}
\newtheorem{remark}[thm]{Remark}

\newcommand{\mean}{\mathbb{E}}

\newcommand{\R}{\mathbb{R}}

\newcommand{\fihc}{\mathcal{F}}
\newcommand{\popprob}{\mathbb{P}}

\renewcommand{\leq}{\leqslant}
\renewcommand{\geq}{\geqslant}

\DeclareMathOperator{\Fill}{Fill}

\pgfplotsset{compat=1.18}

\begin{document}

\title{Availability of Perfect Decomposition in Statistical Linkage Learning for Unitation-based Function Concatenations \tnoteref{fund}}

\author[1]{Michal Prusik}
%\ead{michal.prusik@pwr.edu.pl}

 \author[1]{Bartosz Frej \corref{cor}}
\ead{bartosz.frej@pwr.edu.pl}

 \author[3]{Michal W. Przewozniczek}

 \affiliation[1]{
 	organization={Faculty of Pure and Applied Mathematics, Wroclaw University of Science and Technology},
 	city={Wroclaw},
 	country={Poland} 
 }

 \affiliation[3]{
 	organization={Departament of Systems and Computer Networks}, Wroclaw University of Science and Technology,
 	city={Wroclaw},
 	country={Poland} 
 }
%\ead{michal.przewozniczek@pwr.edu.pl}

\cortext[cor]{Corresponding author at: Wroc{\l}aw University of Science and Technology, Faculty of Pure and Applied Mathematics, Wybrze\.{z}e Wyspia\'{n}skiego 27, 50-370 Wroc{\l}aw, Poland}
\tnotetext[fund]{This work was supported by the Polish National Science Centre (NCN) under Grant 2022/45/B/ST6/04150.}

\begin{abstract}
Statistical Linkage Learning (SLL) is a part of many state-of-the-art optimizers. The purpose of SLL is to discover variable interdependencies. It has been shown that the effectiveness of SLL-using optimizers is highly dependent on the quality of SLL-based problem decomposition. Thus, understanding what kind of problems are hard or easy to decompose by SLL is important for practice. In this work, we analytically estimate the size of a population sufficient for obtaining a perfect decomposition in case of concatenations of certain unitation-based functions. The experimental study confirms the accuracy of the proposed estimate. Finally, using the proposed estimate, we identify those problem types that may be considered hard for SLL-using optimizers.
\end{abstract}

\begin{keyword}
evolutionary algorithm \sep linkage learning \sep optimization \sep problem decomposition \sep dependency structure matrix \sep entropy
\end{keyword}

\maketitle

\newcommand{\theordep}{\widetilde{Q}}

\section{Introduction}

\subsection{Background}
\label{sec:intro}

Evolutionary Algorithms (EAs) process the population of solutions using various operators, e.g. mixing (crossover), mutation, perturbation, and others~\cite{ltga,ilsDLED,nbDiscrGomea}. EAs are state-of-the-art optimizers for many real-world problems~\cite{nonBinaryELL}. Thus, improving their effectiveness and efficiency is important for practical purposes. Many EAs employ knowledge of variable interdependencies to improve their effectiveness. This applies to various problems and solution spaces, e.g., continuous search spaces~\cite{ePX,dg,dg2,irrg}, permutation-based problems~\cite{rkEda,hybridLLperm,ltgaPopulationSizing} and binary~\cite{pxForBinary} or non-binary discrete search spaces~\cite{nonBinaryELL,nbDiscrGomea}. Some research papers directly state that knowledge about variable dependencies should be utilized whenever possible~\cite{whitley2019}.\par

In the so-called gray-box optimization \cite{whitley2019}, all variable dependencies are known a priori, which allows the construction of EA-dedicated operators \cite{pxForBinary}. However, in black-box optimization, dependencies between variables are unknown and need to be discovered. To this end, either the optimization process may be divided into two phases (discovering the dependencies and then searching for high-quality solutions) \cite{dg,dg2,irrg} or variable dependencies are discovered on the run, i.e., the process of dependency discovery is mixed with the high-quality solution search \cite{FIHCwLL,ilsDLED,ltga,dsmga2}. Some studies point out that discovering the dependencies on the run seems to be a better and more flexible choice \cite{linkLearningDetermined}. \par

To find dependencies between variables (\emph{genes}), one may use different strategies, known under the common name \textit{linkage learning} (LL).
It was noticed and verified experimentally that the effectiveness of LL-using EAs significantly depends on their capability of precise discovery of variable dependencies \cite{ltga,dgga,pxForBinary}. 
In this work, we focus on Statistical Linkage Learning (SLL) employed by many state-of-the-art optimizers, e.g., Linkage Tree Gene-pool Optimal Mixing Evolutionary Algorithm (LT-GOMEA) \cite{ltgaPopulationSizing} and Parameter-less Population Pyramid (P3) \cite{P3Original}. In SLL, the dependencies are discovered by performing statistical analysis on the population of solutions that were subject to optimization in the prior optimizer iterations. The information gained from this process is then stored for further use and is treated as the ultimate source for retrieving knowledge on the dependency structure of the problem (see Section~\ref{sec:basicNotions} for precise statements).\par

Deceptive functions \cite{decFunc,decBimodalOld} are tools frequently used to construct hard-to-solve artificial problems to verify optimizers' effectiveness. For binary problems, their values often depend on the so-called \textit{unitation}, i.e., the sum of binary variables. Note that that SLL-using optimizers are in the group of EAs that are the most effective in solving problems built from deceptive functions \cite{ltga,P3Original,dsmga2,3lo,FIHCwLL}. These, in turn, are employed to understand the pros and cons of a given optimizer, which, for SLL-using optimizers, includes the assessment of the quality of information gained from SLL. To the best of our knowledge, this issue was initially addressed in \cite{linkageQuality}, where the size of a population supporting perfect decomposition was given only in the case of standard deceptive functions. However, a similar estimation for other, more sophisticated deceptive functions is unavailable. Therefore, the main objective of this work is to extend the results of \cite{linkageQuality} and to propose an approach that yields an assessment of how difficult the problem is for optimizers using SLL.\par

\subsection{Related work}
\label{sec:rw}

\subsubsection*{Deceptive functions and the dependency structure}
\label{sec:rw:deps}

Deceptive functions were proposed in \cite{decFunc,decBimodalOld} to construct hard-to-solve optimization problems. To present the ideas that stand behind them, let us first analyze a Onemax problem that is easy to optimize. In Onemax, the fitness is the sum of binary values that are the entry parameters of a given function. In this case, each variable can be optimised separately (simply set to one) to find an optimal solution. Thus, all variables may be considered independent.\par

Let us now consider the standard deceptive function of order $k$  defined as:

\begin{equation}
	\label{eq:trap}
	\mathit{trap}_k(u) = 
	\begin{cases}
		k - u - 1,& u < k \\
		k, & u = k\\
	\end{cases},
\end{equation}
where $k$ is the function size.

Every standard deceptive trap function has one local maximum (for $u=0$) and one global maximum (for $u=k$). In this problem, all variables are considered directly dependent, as the set of variables cannot be decomposed into subsets for which optimization may be performed separately. The next level of difficulty is represented by bimodal deceptive functions having two global maxima and a suboptimal maximum in between or noised bimodal functions with a larger number of local maxima. The experimental results confirm that the problems built from bimodal functions, all the more for the noised bimodal functions, are hard to optimize for SLL-using optimizers \cite{3lo,dgga,FIHCwLL} \par

Deceptive functions are concatenated to form blocks of directly dependent genes. Such blocks may overlap, possibly yielding a complicated graph of dependencies. For instance, consider
\[f_{overlap}(x_1,x_2,x_3,x_4)=\mathit{trap}_3(u(x_1,x_2,x_3)) + \mathit{trap}_3(u(x_2,x_3,x_4)).
\]
Variables $x_1$ and $x_4$ are not directly dependent on each other because $x_4$ does not affect partial optimization performed on a subset $x_1,x_2,x_3$ (and \textit{vice versa}). However, they are dependent indirectly because they are both directly dependent on $x_2$. The analysis of overlapping problems seems much more complex than the study of problems that can be decomposed into disjoint blocks, which do not interfere with each other.\par

\subsubsection*{Statistical linkage learning and its operators}
\label{sec:rw:sll}

SLL is based on the construction of the dependency structure matrix (DSM), derived from the organization theory~\cite{dsmga2}. DSM is a square matrix, which stores the discovered variable dependencies, and its entries $d_{i,j}$ refer to the strength of dependency between $i^{th}$ and $j^{th}$ variable (see Def.~\ref{def:inf} in the current paper). DSM is employed to obtain clusters of dependent genes. Frequently, these clusters are obtained by the Linkage Tree (LT) construction in the following way \cite{ltga,P3Original,3lo}. First, for each gene, a single leaf that contains only this gene is created. Then, the most dependent LT nodes are joined with respect to DSM until all nodes are joined. Except for the LT root, each LT node may serve as a mixing mask that groups the dependent genes.\par

Many state-of-the-art SLL-based EAs use the Optimal Mixing (OM) operator \cite{P3Original,ltga,dled,3lo,FIHCwLL}. The entry variables of OM are the mixing mask and the \textit{source} and \textit{donor} individuals. Genes marked by the mask are copied from the donor to the source individual. The modification is preserved if the source's fitness does not decrease. Otherwise, it is rejected. Note that some recent works propose and analyze more sophisticated strategies for modification preservation/rejection \cite{slideOrNot}.\par

In \cite{linkageQuality} the $\Fill$ measure was proposed to measure the DSM quality. It is defined as follows:

\begin{equation}
	\label{eq:linkQualityFill}
	\Fill(i, M) = \frac{ TrueDep(i,M,BlockSize(i))}{BlockSize(i)},
\end{equation}
where $i$ is the position in the genotype and $M$ is a DSM, $BlockSize(i)$ is the number of genes that depend on the $i^{th}$ gene, and $TrueDep(i,M,k)$ is the number of genes that are correctly pointed by the matrix $M$ to be among $k$ genes, most dependent on the $i^{th}$ gene.
If for a given $i^{th}$ gene, DSM $M$ marks as the most dependent of those genes that are truly dependent on it, then $\Fill(i,M)=1$ that corresponds to the perfect linkage. Clearly, if among $BlockSize(i)$ genes pointed out by $M$ as the most dependent on the $i^{th}$ gene, there are no genes that are truly dependent on it, then $\Fill(i,M)=0$, which corresponds to the linkage of the worst quality.

\subsubsection*{SLL-using optimizers}
\label{sec:rw:opts}

SLL and OM are employed by LT-GOMEA, which was originally proposed to optimize permutation-based problems \cite{ltgaPopulationSizing} but can be applied to binary domains as well \cite{3lo,FIHCwLL}. In LT-GOMEA, the typical GA-like operators, i.e., crossover, mutation, and selection, are replaced with OM. Additionally, LT-GOMEA employs the population-sizing scheme \cite{HarikPopSizing}, which makes it parameter-less. During its run, it creates populations of increasing size (some are deleted if they are considered useless). Each population constructs its own DSM and uses it for OM.\par

In P3 \cite{P3Original} during each iteration, a new individual is created randomly and then it is optimized by the First Improvement Hill Climber (FIHC) \cite{P3Original,3lo,FIHCwLL}---a local search optimizer presented in Pseudocode \ref{alg:fihc} (see section \ref{sec:symm_probl_analys}). During its procedure, all genes are flipped in a random order. If a given flip operation improves fitness, then it is preserved or reverted otherwise. In each FIHC iteration, each gene is considered (flipped) once. The iterations are executed until no modification is preserved.\par

The population in P3 is organized in a pyramid-like manner. The first level contains only individuals that were optimized by FIHC. During each iteration, a new individual (denoted as \textit{climber}) is mixed with the subsequent pyramid levels using OM. If, after mixing with a given pyramid level, the fitness of the climber is improved, then its copy is added to the next level of the pyramid (if such a level does not exist, then it is created and initialized with this single individual). In general, the solutions of higher quality are located on the higher levels. Each level has its separate DSM that is used for mixing with its individuals.\par

A standard version of LT-GOMEA does not use local search to initialize individuals, but such LT-GOMEA versions were considered recently in \cite{FIHCwLL}. The locally optimal solution seems particularly useful for optimization \cite{locOptAreGood}. The analysis of the expected population size of FIHC-initialized individuals was helpful in explaining why problems built from the standard deceptive function are easy to solve for SLL-using optimizers \cite{linkageQuality}.

\subsubsection*{Empirical linkage learning}
To discover variable dependencies in black-box optimization, many of the state-of-the-art optimizers perturb solutions and perform a dependency check \cite{irrg,ilsDLED,dg,dg2}. Frequently, these checks refer to non-linearity or non-monotonicity checks \cite{goldMuneDepChecks}. The advantage of such \textit{empirical linkage learning} techniques (ELL) is that they never discover \textit{false linkage}, i.e., they never report two independent variables as dependent \cite{3lo}. However, their disadvantage is they usually do not measure the strength of dependency. Note that the research on enhancing ELL with the dependency strength measurement is promising \cite{cannibal} but remains in its early stages \cite{GAwLL}.\par

\subsection{Article contents}
\label{sec:content}

Many state-of-the-art SLL-using optimizers use local search during the initialization process of new solutions~\cite{P3Original,dsmga2,3lo,FIHCwLL}. These solutions are later on the subject of the SLL linkage discovery procedure. Some works directly conclude that locally optimal solutions have features that make them valuable for a further optimization process~\cite{locOptAreGood}. Therefore, similarly to \cite{linkageQuality}, we estimate the population size of locally optimal solutions that is enough to obtain a perfect SLL-based decomposition.\par

To propose the aforementioned estimation, we use the geometry of the space of probability distributions and some probabilistic and combinatorial tools, in particular the Chernoff bounds \cite{Chernoff}, we propose a quantitative study of parameters of SLL techniques when significantly more complicated functions are considered. The main motivation was to consider the bimodal functions of unitation, which seem to be hard to decompose by SLL \cite{3lo}. However, the proposed approach is significantly more general, e.g. it covers the case of symmetric noised bimodal functions. Further analysis allows for the identification of those problems that may be impossible to decompose by SLL under some conditions.\par

The rest of this paper is organized as follows. In Section \ref{sec:basicNotions}, we present probabilistic and geometric definitions and ideas, which we use in our arguments. Section \ref{sec:symm_probl_analys} contains the formulation and the proof of our main result---the estimate of the size of the population sufficient for a perfect decomposition in a certain class of problems. In Section \ref{sec:bimod_and_revert}  we apply our theorem to two example problems.  Section \ref{sec:GeneralCase} presents a discussion of phenomena that appear in the analysis of more general cases. In Section \ref{sec:exp}, we report and analyze the results of the experiments, comparing them to the theoretical background. Finally, the last section concludes this work and defines the most promising future work directions.

\section{Basic notions}
\label{sec:basicNotions}

Consider a population of $s$ solutions, i.e., $s$~binary sequences $(x_1,...,x_n)$ of fixed length. We treat each gene as a binary random variable with empirical distribution $P_i=(p_i,1-p_i)$ obtained from a given population, namely, $p_i$ is the frequency of zeros at $i$th coordinate and $1-p_i$ is the frequency of ones. For instance, the distribution of the first gene is the vector $(p_1,1-p_1)=(\frac{s_0}{s},\frac{s-s_0}{s})$, where $s_0$ denotes the number of solutions in the population with $x_1=0$. 
Similarly, we will define empirical distributions of pairs of genes.
To keep the notation short and simple, we will prefer to write binary pairs in the form $00$ rather than $(0,0)$.
Then, we let $P_{ij}=(p_{ij}(00),p_{ij}(01),p_{ij}(10),p_{ij}(11))$, where $p_{ij}(00)
=\frac{s_{00}}{s}$ for $s_{00}$ being the number of members of the population with $00$ on genes $i$ and $j$, etc.

The following notions are essential for our consideration.
Let $X$ and $Y$ be random variables with {finite }ranges $\mathcal{X}$, $\mathcal{Y}$ and distributions $p_X$, $p_Y$, respectively. Namely, $p_X(x)$ denotes the probability that the varaiable $X$ takes the value~$x$.
Similarly, let $p_{XY}(x,y)$ be the joint probability of $X$ and $Y$ , i.e., the probability that simulataneously $X=x$ and $Y=y$.
\begin{defn} \label{def:inf}
	\begin{enumerate}
		\item The \emph{entropy} of $X$ is given by 
		\[
		H(X) = -\sum_{x\in \mathcal{X}} p_X(x) \log p_X(x).
		\]
		\item The \emph{joint entropy} of $X$ an $Y$:
		\[
		H(X,Y) = -\sum_{x\in\mathcal{X}} \sum_{y\in\mathcal{Y}} p_{XY}(x,y)\log p_{XY}(x,y)
		\]
		\item The \emph{mutual information} of $X$ and $Y$:
		\[
		I(X,Y) = \sum_{x\in\mathcal{X}} \sum_{y\in\mathcal{Y}} p_{XY}(x,y) \log\frac{p_{XY}(x,y)}{p_X(x)p_Y(y)}.
		\]
		\item The \emph{distance between random variables} $X$, $Y$ is given by by
		\[
		D(X,Y) = 1-\frac{I(X,Y)}{H(X,Y)}.
		\]
		\item For a finite sequence $(X_i)$, $i=1,...,n$, of random variables the \emph{dependency structure matrix (DSM)} is the matrix with entries $D_{i,j} = D(X_i,X_j)$.
	\end{enumerate}
\end{defn}
\begin{remark}
	The distance $D(X,Y)$ is equal to one if and only if $X$ and $Y$ are stochastically independent, while it is zero only if $X$ and $Y$ strictly determine each other, i.e., $Y=F(X)$ for a certain invertible function $F$.  
\end{remark}

\begin{remark}
	In our setup, we will prefer to write the entropy as $H(P)$---a function of a probability vector defining the distribution of a respective random variable. We do the same for joint entropy or mutual information of two genes, so the above definitions take the following forms: 
	\begin{gather*}
		% entropy
		H(P_i) =  -p_i\log p_i - (1-p_i)\log(1-p_i)\\
		% joint entropy
		H(P_{ij}) =  -\sum_{b_1,b_2\in\{0,1\}} p_{ij}(b_1b_2)\log p_{ij}(b_1b_2)\\
		% mutualinformation
		I(P_{ij})= \sum_{b_1\in\{0,1\}} \sum_{b_2\in\{0,1\}} p_{ij}(b_1b_2)\log\frac{p_{ij}(b_1b_2)}{p_i(b_1)p_j(b_2)}.
	\end{gather*}
	It is well known that 
	\begin{equation}    \label{eq:mut_inf}
		I(P_{ij})=H(P_i)+H(P_j)-H(P_{ij}).
	\end{equation}
\end{remark}

We emphasize that $H(P_i)$, $H(P_{ij})$ and $I(P_{ij})$ are counted with respect to probabilities obtained as frequencies of symbols or pairs of symbols in the population. Thus, they are random variables---functions of probability vectors coming from random populations. 
In the following, we are going to compare such empirical distributions with theoretical distributions using their relative entropy.
\begin{defn} The \emph{relative entropy} (or \emph{Kullback-Leibler divergence}),  of two-valued distributions $(p,1-p)$, $(q,1-q)$ is given by:
	\[
	H_q(p) = p\log\frac{p}{q} + (1-p)\log\frac{1-p}{1-q}.
	\]
\end{defn}
By a straightforward calculations one proves the following properties of the relative entropy.
\begin{lem} \label{lem:rel_entropy}
	\begin{enumerate}
		\item $H_q(p)=H_{1-q}(1-p)$ \label{lem:rel_entropy:symm}
		\item For a fixed $q$ the function $H_q(p)$ is monotone decreasing for $p<q$ and increasing for $p>q$, \label{lem:rel_entropy:monot}
		\item $H_q(p) \leq H_q(\frac12-p)$ for $p,q\in[\frac14,\frac12]$. \label{lem:rel_entropy:ineq}
	\end{enumerate}
\end{lem}

\begin{figure*}[ht]
	\centering    \includegraphics[width=0.9\linewidth] {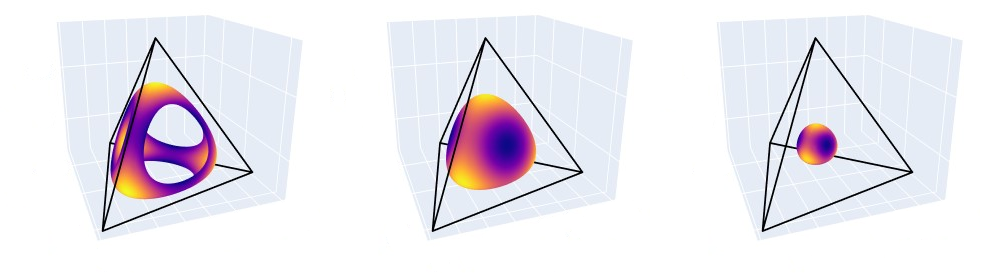}
	\caption{Level sets of entropy function for three values: $\log{2.6}$, $\log{3}$ and $\log{3.8}$, increasing from left to right.} 
	\label{fig:level_sets}
\end{figure*}

The set of all probability vectors of length 4 is a three-dimensional simplex~$\mathcal{S}$ (a tetrahedron) in $\R^4$, with four extremal points (vertices) having 1 at a single non-zero coordinate. The entries of every such probability vector are called barycentric coordinates of the vector regarded as a point in the simplex, i.e., they represent each probability vector as a convex combination of extremal points. The boundary of the simplex consists of four faces being itself 2-dimensional simplices (i.e., equilateral triangles). Each such face consists of vectors whose one fixed coordinate is equal to zero.  It is important that the entropy $H(P)$ is a nonnegative, continuous and \textit{concave} function defined on the simplex. Our arguments involve analysis of level sets of the entropy function on $\mathcal{S}$, i.e., sets defined as $\mathcal{L}_h=\{P\in\mathcal{S}: H(P)=h\}$ for a fixed $h\in[0,\log 4]$
(see Figure \ref{fig:level_sets} for three examples of the level sets of the entropy in the simplex $\mathcal S$).
It follows easily from continuity and concavity of entropy that the sets $\{P\in\mathcal{S}: H(P)\geq h\}$ bounded by $\mathcal{L}_h$ are closed and convex.

Let $\widetilde{H}$ be the function defined on $\{(p_1,p_2,p_3,p_4): p_1,...,p_4\geq0 \}$  by
\[
\widetilde{H}(p_1,p_2,p_3,p_4) = -\sum_{i=1}^4 p_i \log p_i.
\]
Then $\widetilde{H}(P)=H(P)$ for $P\in\mathcal{S}$.
Let $\nabla \widetilde{H}$ denote the gradient of $\widetilde{H}$ and $\nabla{H}_\mathcal{S}$ be its orthogonal projection onto the hyperplane containing $\mathcal{S}$. Note that this hyperplane is unambiguously defined by the four vertices of the simplex or by demanding that coordinates of the point belonging to the hyperplane sum to 1.
\begin{lem} \label{lem:dot_prod_with_gradient}
	Let $P$ and $Q$ be probability vectors. If the dot product $(P-Q)\cdot \nabla \widetilde{H}(Q)$ is less than 0, then $H(P) \leq H(Q)$.
\end{lem}
\begin{proof}
	If $H(P) > H(Q)$ then by concavity of the entropy function on~$\mathcal{S}$
	\[
	\widetilde{H}(Q+t(P-Q)) = \widetilde{H}((1-t)Q+tP) \geq (1-t)\widetilde{H}(Q)+t \widetilde{H}(P) > \widetilde{H}(Q)
	\]
	for each $t\in(0,1)$. Hence $(P-Q)\cdot \nabla \widetilde{H}(Q)$ is nonnegative as the directional derivative of $\widetilde{H}$ along $(P-Q)$.
\end{proof}
\begin{remark}
	Note that $(P-Q)\cdot \nabla \widetilde{H}(Q)=(P-Q)\cdot \nabla{H}_\mathcal{S}(Q)$ for $P,Q\in\mathcal{S}$, because $\nabla{\widetilde{H}}(Q)-\nabla{H}_\mathcal{S}(Q)$ is orthogonal to $P-Q$.
\end{remark}
\begin{lem} \label{lem:parallel_gradient}
	For each $P=(t,\frac12-t,\frac12-t,t)$ with $t\in(\frac14,\frac12)$ there is $k>0$ such that $\nabla H_\mathcal{S}(P)=(-k,k,k,-k)$.
\end{lem}
\begin{proof}
	The vector $(1,1,1,1)$ is normal to the hyperplane containing $\mathcal{S}$
	, hence for $P=(t,\frac12-t,\frac12-t,t)$ we have $\nabla{H}_\mathcal{S}(P)=\nabla{\widetilde{H}}(P)-(s,s,s,s)$, where~$s$ is suitably chosen. 
	This yields the equation
	\[
	\big(\nabla{\widetilde{H}}(P)-(s,s,s,s)\big)\cdot(1,1,1,1)=-2\log t - 2\log\big(\frac12-t\big)-4 - 4s = 0,
	\]
	hence $s=-\frac{\log t + \log\big(\frac12-t\big)}{2} - 1$ and, finally, $\nabla{H}_\mathcal{S}(P) = (-k,k,k,-k)$ for
	\[
	k=\frac{\log t - \log\big(\frac12-t\big)}{2}.
	\]
\end{proof}
%
%
%==========================================================================
%
%

\section{Theoretical analysis of symmetric problems}
\label{sec:symm_probl_analys}

\subsection{Discovering dependencies}

To get a grip on the impact of dependencies onto the optimization process, we perform local search using the First Improvement Hill Climber. Whenever we refer to the results of FIHC procedure applied to the whole population, we assume that the order of climbing (i.e., visiting the coordinates) was chosen individually and independently for each member of the population, by means of the uniform distribution. In a nutshell, the decomposition runs in the following steps:

\begin{enumerate}
	\item the population of $s$ individuals $x=(x_1,...,x_n)$ is sampled and each individual undergoes FIHC optimization,
	\item based on this modified population the DSM  is constructed,
	\item genes are grouped in blocks according to DSM entries.
\end{enumerate}

\begin{algorithm}
	\caption{First Improvement Hill Climber}
	\begin{algorithmic}[1]
		\Function{RunForLevel}{$solution$}
		\State  $optSolution \gets solution$
		\State $geneOrder \gets $ GenerateRandomGeneOrder(size($optSolution$));
		\Repeat
		\State $modified \gets false$;
		\For{\textbf{each} $gene$ in $geneOrder$} 
		\State $fitness \gets$ Fitness($optSolution$);
		\State $optSolution[gene] \gets \neg optSolution[gene]$;
		\State $fitnessNew \gets$ Fitness($optSolution$);
		\If {$fitnessNew > fitness$}
		\State $modified \gets true$;
		\Else
		\State $optSolution[gene] \gets \neg optSolution[gene]$;
		\EndIf
		\EndFor 
		\Until{$modified = true$};
		\State \Return {$optSolution$}
		\EndFunction
	\end{algorithmic}
	\label{alg:fihc}
\end{algorithm}
To understand the idea behind the first, preparatory step, note that the initial population is chosen randomly from the uniform distribution, so it reveals no dependencies between genes. However, such dependencies influence the process of optimization. The configurations of symbols which occur in highly fitted solutions have tendency to be more frequent, so by observing frequencies of certain configurations one may draw conclusions on dependencies. 
%Similar effect may be obtained by a selection procedure, i.e., by choosing a bigger population and restricting to a fraction of best fitted solutions. This procedure would depend on two parameters (the size of the initial population and the size of the fraction of best individuals) instead of one. It would also introduce uncontrolled randomness coming from uncertainty concerning presence of well fitted solutions in the initial population. The cost of the FIHC procedure involves $sn$ evaluations of the fitness function as opposed to $s\gamma$ for selection procedure, where $\frac1\gamma$ is the desired fraction of best fitted solution, so we choose the one which gives us better control over the process.

\begin{defn} 
	We say that the DSM provides a \emph{perfect decomposition} of the problem if all entries corresponding to distances between dependent variables are smaller than the entries given by independent pairs.
\end{defn}
\begin{remark}
	In the above definition we want to specify which matrices contain information sufficient to distinguish dependent pairs of genes from independent ones. It must be emphasized that the decomposition into blocks of dependency may be obtained via different procedures.
	However, if the decomposition is done by the Linkage Tree construction, then dependencies between clusters of genes are determined by such pairwise dependencies.
\end{remark}
Note that to ensure that $D(X,Y)<D(X,Z)$ it is enough that $ H(X,Y) < H(X,Z)$ and $I(X,Y) \geq I(X,Z)$.
Willing to employ the aforementioned notation, 
we rewrite it as
\begin{equation}	\label{cond:inf_joint}
	H(P_{i,j}) < H(P_{i,m}) \qquad \mathrm{and}\qquad I(P_{i,j})\geq I(P_{i,m}).
\end{equation}
By \eqref{eq:mut_inf}, the second inequality may be rewritten as 
\[
H(P_j) - H(P_{i,j}) \geq H(P_m) - H(P_{i,m}).
\]

\subsection{Properties of symmetric problems}
Below we will consider binary problems of length $n=rk$, 
which are concatenations of functions $g\circ u$, i.e.,
\begin{equation}    \label{eq:form}
	f(x)=\sum_{i=0}^{r-1} g\circ u(x_{ki+1},...,x_{k(i+1)}),    
\end{equation}
where $u:\{0,1\}^k\to\{0,...,k\}$, $u(x)=\sum_{i=1}^k x_i$, is the unitation function and $g:\{0,...,k\}\to\R$ is symmetric in the sense that $g(i)=g(k-i)$ for each $i=0,...,k$.  From now on, by \emph{blocks} we will mean tuples of genes $(x_{ki+1},...,x_{k(i+1)})$ occurring as arguments of a single partial function. 
The main facilitation of considering non-overlapping blocks lies in the fact that a hill-climbing optimizer works separately in each of concatenated blocks, and the final value of the gene $x_i$ is not affected by blocks that do not contain~$x_i$. 
Therefore, we say that $x_i$ and $x_j$ are independent if they lie in disjoint blocks. Otherwise, they depend on each other. 

In the following we will also consider genes of a single solution as random variables, whose distributions are determined by the way we initialize populations. As it was already described, each population (a base for the empirical distribution) is sampled from various $s$ element populations with probability $1/2^{ns}$ and then each of its members is optimized by FIHC proceeding in random order, selected with equal probability from among $n!$ various permutations. The probability measure, which is the outcome of this procedure, will be referred to as \emph{the theoretical distribution} of a gene in a solution. Note that in each solution the distribution of the $i$th gene is the same, so we can just speak about the theoretical distribution of the $i$th gene. The same argument can be applied to pairs of genes.
Also, the empirical distributions $p_i(b)$ and $p_{ij}(b_1b_2)$, where $b,b_1,b_2\in\{0,1\}$, are random variables on the domain of all possible populations.

The independence of genes, understood as belonging to disjoint blocks, implies their stochastic independence with regard to their theoretical distributions. However, we will see that these two notions are not equivalent. Moreover, since the FIHC procedure acts in a random order and all genes have an equal impact on the unitation function, the distribution of a gene does not depend on its position in the respective block. In other words, the distribution of a single gene, as well as the distribution of a pair of genes, is common for all genes (or all pairs of genes) in the same block. The same properties, together with the symmetry of $g$, also assure that the probability of seeing 1 at a fixed gene is equal to $\frac12$.

\begin{figure}[ht]
	\centering
	\includegraphics[width=0.7\linewidth] {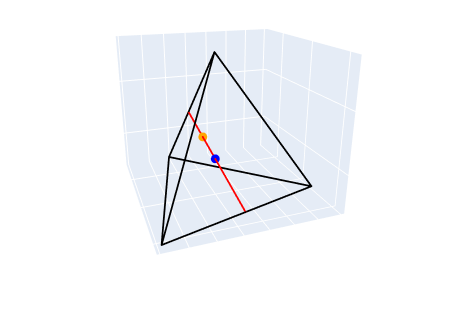}
	\caption{The interval $\mathcal I$ with the center of the simplex $(\frac14,\frac14,\frac14,\frac14)$ (blue point) and the distribution of dependent pairs $\theordep$ (orange point).} 
	\label{fig:interval}
\end{figure}

The sequence of values of the fixed gene throughout a population forms a Bernoulli process. Knowing that the expectation of the binomial distribution $B(s,p)$ for the number of successes in $s$ trials is equal to $sp$ we obtain $\mean p_i= \frac12$.
For similar reasons, if genes $x_i$ and $x_m$ lie in separate blocks, zeros and ones appear independently and then $\mean p_{im}(b_1b_2)= \frac14$.
On the other hand, the symmetry property implies that for a pair of genes $x_ix_j$ sharing the same block, the theoretical probability of seeing $00$ is equal to the probability of $11$, and the probability of $01$ is the same as of $10$. Hence, the theoretical distribution of the dependent pair has the form $(q,\frac12-q,\frac12-q,q)$, $q\in[0,\frac12]$, for each such pair $x_ix_j$, so, geometrically, it lies in the three-dimensional simplex on the interval $\mathcal I$ connecting points $(\frac12,0,0,\frac12)$ and $(0,\frac12,\frac12,0)$ (see Figure \ref{fig:interval}). 

\subsection{Main theorem}
From now on, we will denote the theoretical probability distribution of a dependent pair by $\theordep=(\tilde q,\frac12-\tilde q,\frac12-\tilde q,\tilde q)$. Because of this symmetry and the fact that all arguments depend exclusively on properties of the simplex of distributions, where the labelling of vertices is not relevant, we may and will assume that $\tilde q \geq \frac14$. More precisely, if $p_1$ is the probability of $00$ and $p_2$ is the probability of $01$, then we take $\tilde q = \max\{p_1,p_2\}$ and reorder the vector $(p_1,p_2,p_2,p_1)$, so that the first and the last entries are equal to $\tilde q$ (possibly, $\tilde q$ stands now for probability $p_2$ of $01$ and $10$). Note that $\tilde q=\frac14$ yields the distribution of an independent pair. Thus, we will assume strict inequality $\tilde q > \frac14$.

\begin{lem} \label{lem:q_of_p}
	Let $P=(p,\frac12-p,\frac12-p,p)$, where $\frac14 < p < \frac12$. There exists exactly one $q=q(p)$ such that $Q=(q,\frac12-q,\frac12-q,q)$ satisfies $H(P)-2H(Q)+\log 4=0$. Moreover, the function $q(p)$ is continuous, increasing with $p$ and $\frac14 < q(p) < p < \frac12$.
\end{lem}
\begin{proof}
	The existence of $q$ follows from the fact that the function $q\mapsto H(Q)$ is decreasing on $[\frac14,\frac12]$, thus it has an inverse. An inverse of a continuous map of a compact interval is automatically continuous. Also, the equality $H(Q)=\frac{H(P)+\log4}{2}$ implies that if $p$ increases then the right hand side decreases, so $q(p)$ must increase. Since $H(Q)>\frac{H(P)+H(P)}{2}=H(P)$ we get $q(p)<p$.
\end{proof}

\begin{thm} \label{thm:main}
	Consider a fitness function of the form $\eqref{eq:form}$. Let $\theordep=(\tilde q,\frac12-\tilde q,\frac12-\tilde q,\tilde q)$ be the probability distribution of values of any (every) dependent pair of genes $x_ix_j$, where $\tilde q >\frac14$ is the greater of the probabilities of $00$ and $01$.
	Then there is a unique $\rho\in(\frac14,\tilde q)$ such that $H_{2\tilde q}(2\rho)=H_{1/4}(q(\rho))$ and for the population of size
	\[
	s_{min} = \left\lceil \frac1{H_{2\tilde q}(2\rho)}\cdot\log\frac{r\binom{k}{2} + 8\binom{r}{2}k^2}{1-\alpha} \right\rceil
	\]
	the DSM yields perfect decomposition with probability greater than or equal to~$\alpha$.
\end{thm}
\begin{proof}
	Without loss of generality, we will assume that $\tilde q$ is the probability of $00$. 
	By (\ref{lem:rel_entropy:monot}) of the Lemma \ref{lem:rel_entropy}, $H_{2\tilde q}(2p)$ decreases with $p<\tilde q$, while $H_{1/4}(p)$ increases for $p>\frac14$. Note that $\lim_{p\to\frac14} q(p)=\frac14$. Moreover,
	\[
	H_{2\tilde q}(2\tilde q) = 0 < H_{1/4}(q(\tilde q))\qquad \textrm{and}\qquad
	H_{2\tilde q}(\frac12) > 0 = H_{1/4}(\frac14).
	\]
	The first assertion follows by observing that $H_{2\tilde q}(2p)-H_{1/4}(q(p))$ is a decreasing continuous function (so it has the Darboux property), hence attains zero in a unique $\rho\in(\frac14,\tilde q)$.
	
	The main line of the proof is as follows. Intuitively, the empirical frequencies of zeros and ones and frequencies of binary pairs will approximate the theoretical distributions (marginal and joint, respectively). By the continuity of the entropy function, the entropies of  empirical distributions will also be close to the theoretical values. (Note that if two variables are independent their probability distribution is $(\frac14,\frac14,\frac14,\frac14)$, whose entropy is equal to $\log 4$.) The probability that the approximation is sufficiently good can be estimated by the Chernoff bound, where ``sufficiently good'' means ``yielding smaller values of the distance $D$ for dependent variables than for independent ones''. The following argument makes the above idea precise, by  using geometry of the simplex.
	
	Let us fix a dependent pair $x_i, x_j$, and let $x_m$ be independent of them. For $\varepsilon_1$, $\varepsilon_2$, $\varepsilon_3>0$ consider the following system of inequalities:
	\begin{equation}	\label{cond:LLN}
		\begin{cases}
			H(P_{i,j}) \leq H(\theordep) + \varepsilon_1\\
			H(P_{i,m}) \geq \log 4 - \varepsilon_2\\
			H(P_i),H(P_j),H(P_m) \geq \log 2 - \varepsilon_3
		\end{cases}.
	\end{equation}
	Assuming that the above holds, the condition \eqref{cond:inf_joint} follows from the demand that $H(\theordep) + \varepsilon_1 < \log 4 - \varepsilon_2$ and $\log 2 - \varepsilon_3 - H(\theordep) - \varepsilon_1 \geq \log 2 - \log 4 +\varepsilon_2$. 
	The latter condition: 
	\begin{equation}	\label{cond:three_eps}
		\varepsilon_1+\varepsilon_2+\varepsilon_3 \leq \log 4 - H(\theordep)
	\end{equation}
	is obviously stronger than the first one. Therefore, if for some $\varepsilon_1$, $\varepsilon_2$, $\varepsilon_3>0$ for any pair of dependent genes $x_i$, $x_j$ and any gene $x_m$ independent of them the inequalities \eqref{cond:LLN} and \eqref{cond:three_eps} simultaneously hold, then the DSM gives perfect decomposition.
	For the sake of simplicity, let us assume that $\varepsilon_2=\varepsilon_3$. If $H(P_{i,m}) \geq \log 4 - \varepsilon_2$ holds for each pair $i$, $m$ of independent variables then $2\log 2 -H(P_i) - H(P_m) \leq \varepsilon_2$ implying that $H(P_i)\geq\log 2-\varepsilon_2$ for each $i$, hence the last line of \eqref{cond:LLN} can be removed.
	
	\newcommand{\deplike}{T}
	\newcommand{\indeplike}{U}
	Fix two probability vectors $\deplike=(t,\frac12-t,\frac12-t,t)$ and $\indeplike=(u,\frac12-u,\frac12-u,u)$, where $\frac14<t<\tilde q$ and $u=q(t)$. Then $\frac14<u<t<\tilde q$ and
	\[
	H(\deplike) - 2H(\indeplike) + \log 4 =0.
	\]
	To ensure that \eqref{cond:three_eps} is satisfied we define
	\[
	\varepsilon_1= H(\deplike) - H(\theordep) \qquad \textrm{and} \qquad \varepsilon_2 = \log 4-H(\indeplike).
	\]
	Now the first two inequalities of \eqref{cond:LLN} mean that $P_{ij}$ are separated from $P_{im}$ by two level sets: $H(P) = H(\deplike)$ and $H(P) = H(\indeplike)$. Consider the inequalities $\frac12-u\leq p_{im}(b_1b_2)\leq u$ for each $b_1,b_2$. Geometrically, they mean that $P_{im}$ lies in an octahedron centered in  $(\frac14,\frac14,\frac14,\frac14)$, whose vertices are common points of the level set $H(P)=H(\indeplike)$ with the intervals connecting centers of opposite edges of $\mathcal{S}$. As stated before, the set $\{P\in\mathcal S: H(P)\geq H(\indeplike)\}$ is convex, so it contains the whole octahedron. 
	On the other hand, by Lemmas \ref{lem:dot_prod_with_gradient} and \ref{lem:parallel_gradient} the condition $H(P) \leq H(\deplike)$ is guaranteed by $(P-\deplike)\cdot (-1,1,1,-1) < 0$. In the language of geometry, this in turn characterizes the set of points separated from the center by the plane tangent to the boundary of the set $\{P\in\mathcal S: H(P)\leq H(\deplike)\}$ in $\deplike$ (see Figure \ref{fig:satelite}).  After simple calculations the latter reduces to $p_{ij}(00)+p_{ij}(11)> 2t$.
	\begin{figure}[ht]
		\centering
		\includegraphics[width=0.9\columnwidth]{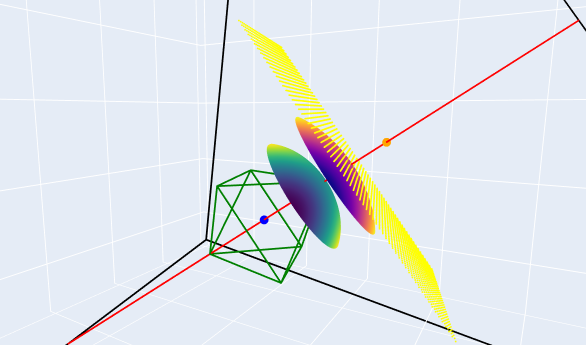}
		\caption{The octahedron contained in $\{P:H(P)\geq H(\indeplike)\}$ and the plane tangent to the level set $H(P)= H(\deplike)$.}
		\label{fig:satelite} 
	\end{figure}
	Concluding, we see that \eqref{cond:LLN} (and \eqref{cond:three_eps}) is implied by
	\begin{equation}    \label{cond:final}
		\begin{cases}
			\frac12-u \leqslant p_{im}(b_1b_2) \leqslant u \qquad \forall\ b_1,b_2\in\{0,1\}\\
			p_{ij}(00)+p_{ij}(11)> 2t
		\end{cases}
	\end{equation}
	with $t$ and $u$ as above.

	Recall that $p_{ij}(b_1b_2)$ are random variables whose values are calculated from a population of $s$ independently chosen solutions $x^1,...,x^s$ as $\frac1s \sum_{l=1}^s \omega_l$, where $\omega_l=1$ if $x^l_{i}=b_1$, $x^l_{j}=b_2$ and $\omega_l=0$ otherwise.
	Using the Chernoff bounds, for each pair $i,j$ of genes belonging to a common block (thus having theoretical probability given by coordinates of $\theordep$)  we obtain 
	\begin{multline}    \label{cond:Chernoff_indep}
		\popprob(p_{ij}(00)+p_{ij}(11) \leqslant 2t)
		= \popprob\left(p_{ij}(00)+p_{ij}(11) - 2\tilde q \leqslant 2(t-\tilde q)\right) \\
		\leqslant \exp(-s H_{2\tilde q}(2t)).
	\end{multline}
	Similarly, if $i$ and $m$ belong to separate blocks then
	\begin{equation}    \label{cond:Chernoff_dep1}
		\popprob(p_{im}\big(b_1b_2) > u\big) =
		\popprob\Big(p_{im}(b_1b_2)-\frac14 > u-\frac14\Big)
		\leq \exp(-s H_{\frac14}(u))
	\end{equation}
	and, using \eqref{lem:rel_entropy:symm} and \eqref{lem:rel_entropy:ineq} of Lemma \ref{lem:rel_entropy},
	\begin{multline}    \label{cond:Chernoff_dep2}
		\popprob(p_{im}(b_1b_2) < \frac12-u) 
		= \popprob(p_{im}(b_1b_2)-\frac14<\frac14-u) \\
		\leq \exp(-s H_{3/4}(u+\frac12)) = \exp(-s H_{1/4}(\frac12 - u))\\
		\leq \exp(-s H_{1/4}(u)).
	\end{multline}
	
	The DSM does not provide the decomposition if for some $i,j$ from the same block and $m$ from another block $D(X_i,X_j)>D(X_i,X_m)$. Then for some $i,j,m$ the condition $\eqref{cond:inf_joint}$ fails. By the above reasoning, the probability of such event is smaller than the probability that either inequality in \eqref{cond:final} fails for some $i,j$ in a common block and $m$ in a separate block.
	Since there are $r$ blocks of length $k$, using subadditivity of probability we can bound it by
	\begin{multline*}
		r\binom{k}{2}\exp(-s H_{2\tilde q}(2t))
		+ 8\binom{r}{2}k^2\exp(-s H_{1/4}(u)) \\
		\leq \Big(r\binom{k}{2} + 8\binom{r}{2}k^2\Big)\exp(-s \min\{H_{2\tilde q}(2t), H_{1/4}(u)\}).
	\end{multline*}
	We recall that $u=q(t)$, $H_{2\tilde q}(2\rho) = H_{1/4}(q(\rho))$ and $H_{1/4}(q(t))$  increases, while $H_{2\tilde q}(2t)$ decreases for $t\in(\frac14,\tilde q)$, so taking $t=\rho$ minimizes the right hand side expression.
	In order to find the proper population size for a probability $\alpha$ it is enough to find minimal $s$ such that 
	\[
	\Big(r\binom{k}{2} + 8\binom{r}{2}k^2\Big)\exp(-s H_{2\tilde q}(2\rho)) \leqslant 1-\alpha,
	\]
	which leads to
	\[
	s_{min} = \left\lceil\frac1{H_{2\tilde q}(2\rho)}\cdot\log\frac{r\binom{k}{2} + 8\binom{r}{2}k^2}{1-\alpha}\right\rceil.
	\]
\end{proof}
We remark that it is hard to find the exact value of $\rho$ analytically, so in the forthcoming examples we will use its close approximation found numerically with the bisection method. It gives the desired number with the tolerance $\frac1{2^\gamma}$, requiring $\gamma$ evaluations of the entropy function, hence its impact on the precision 
%(and runtime) 
is negligible.
%------------------------------
%
\section{Two examples of how to apply the estimate}
\label{sec:bimod_and_revert}
\subsection{Bimodal deceptive functions}
Let us consider a concatenation of $r$ bimodal deceptive functions of order $k=2l$, i.e., functions defined by
\[
b(u) = 
\begin{cases}
	l-|u-l|-1 & \textrm{for } 0<u<2l,\\
	l & \textrm{for } u=0 \textrm{ or } u=2l.
\end{cases}
\]

For $x\in\{0,1\}^{2k}$ denote by $\fihc(x)$ a result of a FIHC procedure applied to $x$ in the natural order, optimizing with respect to the above function.
We have
\[
\fihc(x) = 
\begin{cases}
	0^{2l} & \textrm{for } x=*0^{2l-1},\\
	1^{2l} & \textrm{for } x=*1^{2l-1},\\
	\textrm{a block with unitation }l & \textrm{otherwise}, 
\end{cases}
\]
where $*0^{2l-1}$ denotes a concatenation of an arbitrary symbol with the block $0^{2l-1}$ of $(2l-1)$ consecutive zeros (analogously for $*1^{2l-1}$). 
The main goal is to calculate $\theordep$.
Assume that we are randomly choosing an individual $x$ consisting of $2l$ genes and order it by a random permutation $\pi$ for the action of FIHC. In other words, we perform an operation $\fihc_\pi(x)=\pi^{-1}\fihc(\pi x)$.
Note that we are slightly abusing the notation by letting $\pi$ act on $\{0,1\}^{2l}$ in an obvious way: $\pi(x)_i=x_{\pi(i)}$.
Let us calculate the probability that after such a procedure, a fixed pair of genes $x_ix_j$ takes value $01$.  Note that if the unitation of $\fihc(x)$ is 0 or $2l$, then all genes have the same value, so the pair $01$ does not occur.
By the law of total probability, the probability $\popprob\big(\fihc_\pi(x)_{ij}=01\big)$ of seeing $01$ at genes $x_i$ and $x_j$ when $\pi$ and $x$ were chosen randomly is:
\[
\popprob\big(\fihc_\pi(x)_{ij}=01\,|\, u(\fihc_\pi(x))=l\big)\cdot \popprob\big(u(\fihc_\pi(x))=l\big) 
\]
Denote by $M$ the number of all pairs $(x,\pi)$ for which $u(\fihc_\pi(x))=l$ holds. Each such pair is chosen with the same conditional probability $1/M$. We have $\fihc_\pi(x)_{ij}=01$ if $\fihc_\pi(x)=y$ for some $y\in\{0,1\}^{2l}$ with $u(y)=l$ and $y_{ij}=01$. There are $\binom{2l-2}{l-1}$ such solutions. On the other hand, for any permutation $\sigma$ we have
\begin{equation}    \label{eq:permuting_fihc}
	\sigma \fihc_\pi(x) = \sigma\pi^{-1}\fihc(\pi x) = (\pi\sigma^{-1})^{-1}\fihc((\pi\sigma^{-1} \sigma x) = \fihc_{\pi\sigma^{-1}}(\sigma x).
\end{equation}
For $y$ satisfying $u(y)=l$, the sets $A_y=\{(x,\pi): \fihc_\pi(x)=y\}$ are disjoint and have the same number of elements, because $A_{\sigma y}=\{(\sigma x,\pi\sigma^{-1}): (x,\pi)\in A_{y}\}$ by \eqref{eq:permuting_fihc}. There are $\binom{2l}{l}$ individuals with unitation $l$, so $M=\binom{2l}{l} |A_y|$. 
In particular, this means that given $u(\fihc_\pi(x))=l$ the events $\fihc_\pi(x)=y$ are equally probable, namely, have probability $\frac{|A_y|}{M}=\frac1{\binom{2l}{l}}$.
The number of all solutions $y$ with $y_{ij}=01$ is equal to $\binom{2l-2}{l-1}$, hence
\[
\popprob\big(\fihc_\pi(x)_{ij}=01\,|\, u(\fihc_\pi(x))=l\big) = \frac{\binom{2l-2}{l-1}}{\binom{2l}{l}} = \frac{l}{2(2l-1)}
\]
Denote by $\popprob(u(\fihc_\pi(x))=l)\,|\, \pi)$ the probability that the optimizing procedure returned an individual with unitation~$l$ if it was performed in the order given by~$\pi$. 
Clearly, $\popprob(u(\fihc_\pi(x))=l)\,|\, \pi)$ is equal to $\frac{2^{2l}-4}{2^{2l}}$,
as the only four solutions which do not lead to unitation equal to $l$ are those with all genes homogeneously equal to zero or to one, possibly except for the first gene. Recall that the probability of choosing a permutation $\pi$ from the set of all possible permutations of $\{1,...,2l\}$ is equal to $\frac1{(2l)!}$. Hence, again by the law of total probability
\[
\popprob(u(\fihc_\pi(x))=l) = \sum_\pi \frac1{(2l)!}\popprob\big(u(\fihc_\pi(x))=l\,|\, \pi\big)
\]
which is equal to $\popprob\big(u(\fihc(x))=l)\,|\, \pi_0\big)$
for any $\pi_0$, and we obtain
\begin{gather*}
	\tilde q =\popprob\big(\fihc_\pi(x)_{ij}=01\big) = \frac{l(2^{2l-2}-1)}{(2l-1)2^{2l-1}}\\
	\frac12 - \tilde q =\popprob\big(\fihc_\pi(x)_{ij}=00\big) = \frac{2^{2l-2}(l-1)+l}{(2l-1)2^{2l-1}}
\end{gather*}
For $l=2$, i.e., for the problem of four genes, $\theordep=(\frac14,\frac14,\frac14,\frac14)$, so all genes are stochastically independent. This means that the problem may be hard to decompose by means of SLL (at least in the way we use it), as samples for any pair of genes coming from the same block are statistically indistinguishable from a pair of independent genes. Furthermore, though probabilities are different from $\frac14$ for $l\not=2$, they converge to $\frac14$ when $l$ (and $k$) tends to infinity, thus the dependencies are harder to find for larger $l$. 

As an example we will calculate $s_{min}$ for $k=2l=6$ and $r=10$ with $\alpha = 0.1$. By the above formulas we get $\tilde q = 0.28125$ and $\frac12 - \tilde q = 0.21875$. We then approximate $\rho \approx 0.27227$ and, finally, we obtain $s_{min} = 18\,046$.

\subsection{Reverted bimodal deceptive functions}
In the following let us perform similar analysis for a concatenation of functions on $\{0,\ldots,2l\}$ defined by
\[
rb(u)= \begin{cases}|l-u| & \textrm{for } u \not= l,\\
	2l & \textrm{for } u=l.
\end{cases}
\]
which we refer to as reverted bimodal deceptive functions. For a block $x$ of length $2l$ we have
\[
\fihc(x) = \begin{cases}
	0^{2l} & \textrm{if } u(x)<l-1 \\
	& \textrm{ or } u(x)=l-1,\ x_1=1,\\
	1^{2l} & \textrm{if } u(x)>l+1\\
	& \textrm{ or } u(x)=l+1,\ x_1=0,\\
	\textrm{a block with unitation }l & \textrm{otherwise}.
\end{cases}
\]
To calculate the distribution of two dependent genes we proceed as before and the only thing that needs recalculating is
\[
\popprob(u(\fihc_\pi(x))=l)\,|\, \pi)=\frac{\binom{2l}{l}+2\binom{2l-1}{l}}{2^{2l}}=\frac{4\binom{2l-1}{l}}{2^{2l}}.
\]
Finally, we get
\begin{gather*}
	\frac12 - \tilde q = \popprob\big(\fihc_\pi(x)_{ij}=01\big) = \frac{\binom{2l-2}{l-1}}{2^{2l-1}},\\
	\tilde q = \popprob\big(\fihc_\pi(x)_{ij}=00\big) = \frac12-\popprob\big(\fihc(x)_{ij}=01\big).
\end{gather*}
Again, for $l=2$ we obtain stochastic independence, which is of no surprise, as in case of four genes the problem is  essentially the same as in the previous subsection (from the point of view of hill climbing optimizer). For increasing $l$, however, the situation changes because the probability $\popprob\big(\fihc_\pi(x)_{ij}=01\big)$ decreases to zero making longer problems much easier to decompose!

As before, we will calculate $s_{min}$ for $k=2l=6$ and $r=10$ with $\alpha = 0.1$. By the above formulas we get $\tilde q = 0.3125$ and $\frac12 - \tilde q = 0.1875$. We then approximate $\rho \approx 0.29487$ and, finally, we obtain $s_{min} = 4\,496$.

\section{What can be said without assuming symmetry}
\label{sec:GeneralCase}

\subsection{Possible distributions}
\begin{figure}[ht]
	\centering
	
	\begin{subfigure}{0.48\linewidth}
		\centering
		\includegraphics[width=0.98\linewidth] {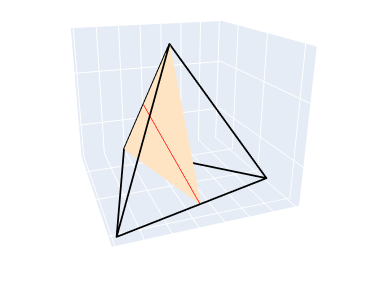}
		\caption{$\mathcal T$ in the simplex.} 
		\label{fig:triangle_3d}
	\end{subfigure}
	\begin{subfigure}{0.48\linewidth}
		\centering
		\includegraphics[width=0.98\linewidth] {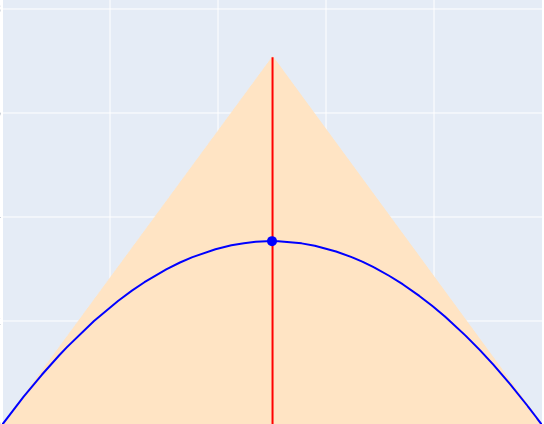}
		\caption{$\mathcal T$ and $\mathcal C$.}
		\label{fig:curva}
	\end{subfigure}

	\caption{The triangle $\mathcal T$ and the curve $\mathcal C$.}
	\label{fig:triangleT}
\end{figure}

{In the current section, we wish to relax the assumption of $g$ being symmetric in the formula \eqref{eq:form}.} Because of the usage of the unitation and uniformly random choice of FIHC orders, all pairs of dependent genes share the same distribution. Here, we will write $p(b_1b_2)$ for the theoretical probability of the pair $b_1b_2$.
In a non-symmetric case we no longer have $p(00)=p(11)$, but the equality $p(01)=p(10)$ {is preserved}, again due to the presence of unitation and varying order of FIHC optimization. Therefore, each pair of dependent genes has the distribution $(p(00),p(01),p(10),p(11))=(q_1,q_2,q_2,q_3)$, which in the simplex is interpreted as a convex combination $q_1(1,0,0,0)+2q_2(0,\frac12,\frac12,0)+q_3(0,0,0,1)$. Any such distribution lies in the triangle $(1,0,0,0), (0,\frac12,\frac12,0), (0,0,0,1)$ (see Figure \ref{fig:triangle_3d}), which will be denoted by $\mathcal T$. The interval $\mathcal{I}$, {the support of} the {theoretic} distributions in the previous case, is the height of the triangle $\mathcal T$. Not every point in the triangle $\mathcal T$  is a distribution of dependent pair of genes in some of considered problems---there are countably many possible distributions, while $\mathcal T$ is an uncountable set. Below, we give formulas for $q_1$, $q_2$, and $q_3$ (depending on the monotonicity of $g$), which can give us some insight into the set of possible distributions, but the exact description of the set is one of the directions for the future work.
\begin{remark}We want to emphasize that in case of our linkage learning procedure, for the problem of the form \eqref{eq:form}, with the symmetry assumption skipped,
	the exact values of the function $g$ are insignificant. The only relevant information affecting the outcome of the optimization process is the decomposition of its domain $\{0,1,\ldots,k\}$ into `monotonicity intervals' . This follows from the fact, that flipping a gene changes the unitation of the individual exactly by one, so FIHC decisions are based only on inequalities between values of $g$ in neighbouring arguments, not the actual magnitudes.
\end{remark}

In the following we assume that there is always a sharp inequality between $g(i)$ and $g(i+1)$. By a local maximum (analogously, minimum) we understand such $i\in\{0,1,\ldots ,k\}$ that $g(i-1)<g(i)$ and $g(i)>g(i+1)$ or just the  appropriate one of these inequalities if $i=0$ or $i=k$. Then, monotonicity of the function $g$ is encoded in two sequences: $MAX_{g}=(k_1,\ldots,k_N)$ and $MIN_g=(l_0,l_1,\ldots,l_N)$, where:
\begin{itemize}
	\item $N$ is the number of local maxima of $g$ on $\{1,\ldots,k-1\}$,
	\item for any $i \in\{1,\ldots,N\}$ we have $l_{i-1}<k_{i}<l_{i}$,
	\item $(k_1,\ldots,k_N)$ are all local maxima of $g$ on $\{1,\ldots,k-1\}$,
	\item $(l_0,l_1,\ldots,l_N)$ are all local minima of $g$ on $\{0,\ldots,k\}$.
\end{itemize}
Additionally, if $l_0>0$ (which means that $g$ has a local maximum at $0$), then we define $k_0=0$. Similarly, if $l_N<k$, we take $k_{N+1}=k$

We are ready to give the formulas for the distribution of the pairs of dependent genes.
\begin{thm}
	\label{distr_general}
	Let $g$ have the monotonicity represented by 
	\[
	MAX_g=(k_1,\ldots,k_N) \qquad \textrm{and} \qquad MIN_g=(l_0,l_1,\ldots,l_N).
	\]
	Let $(q_1,q_2,q_2,q_3)$ be the distribution of the dependent pairs in this model. Then we have
	\begin{eqnarray*}
		q_1 &=& 2^{-k+1}\cdot\sum_{i=1}^N \frac{(k-k_i)(k-k_i-1)}{k(k-1)}\sum_{j=l_{i-1}}^{l_i-1} \binom{k-1}{j} \\
		&+& 2^{-k}\cdot\left(\sum_{j=0}^{l_0-1}\binom{k}{j} + \binom{k-1}{l_0-1}\right) \cdot \chi_{\{l_0>0\}},\\
		q_2 &=& 2^{-k+1}\cdot\sum_{i=1}^N \frac{k_i(k-k_i)}{k(k-1)}\sum_{j=l_{i-1}}^{l_i-1} \binom{k-1}{j},
	\end{eqnarray*}
	where $\chi_{\{l_0>0\}}=
	1$ for $l_0>0$ and 
	$\chi_{\{l_0>0\}}=0$ if $l_0=0$.
\end{thm}
\begin{proof}
	Basically, we will repeat and generalize the reasoning presented in Section \ref{sec:bimod_and_revert}. Take $x\in\{0,1\}^{k}$ and fix two genes $x_v,x_w$.	
	We will start with computing $q_2$, since its formula will not depend on $l_0$. First, let us notice that the situation $x_v=0$, $x_w=1$  may appear after FIHC optimization only if the optimized solution has unitation equal  to $k_i$ for some $i\in\{1,\ldots,N\}$ (i.e., if we are at some maximum which is not the block of zeros and not the block of ones). Thus we have
	\begin{multline*}
		q_2=\popprob(\fihc_\pi(x)_{vw}=01)\\
		=\sum_{i=1}^{N}\popprob(\fihc_\pi(x)_{vw}=01|u(\fihc_\pi(x))=k_i)\popprob(u(\fihc_\pi(x))=k_i).
	\end{multline*}
	By similar arguments as in Section \ref{sec:bimod_and_revert}, the events $\fihc_\pi(x)=y$  are equally probable for all $y\in\{0,1\}^k$ having unitation equal to $k_i$. Hence, to compute the probability $\popprob(\fihc_\pi(x)_{vw}=01|u(\fihc_\pi(x))=k_i)$ we just need to count the number of blocks~$y$ with unitation equal to $k_i$ in which $y_{vw}=01$ and divide it by the number of all such blocks $y$. Therefore we get
	$$\popprob(\fihc_\pi(x)_{vw}=01|u(\fihc_\pi(x))=k_i)=\frac{\binom{k-2}{k_i-1}}{\binom{k}{k_i}}=\frac{k_i(k-k_i)}{k(k-1)}.$$
	To compute $\popprob(u(\fihc_\pi(x))=k_i)$ let us first write it as
	$$\popprob(u(\fihc_\pi(x))=k_i)=\sum_{\pi}\frac1{k!}\popprob(u(\fihc_\pi(x))=k_i|\pi).$$
	For any permutation $\pi$ we get $u(\fihc_\pi(x))=k_i$ whenever one of the following holds:
	\begin{itemize}
		\item $u(x)$ is greater than $l_{i-1}$ and smaller than $l_i$,
		\item $u(x)=l_{i-1}$ and $\pi x_{1}=0$,
		\item $u(x)=l_i$ and $\pi x_{1}=1$.
	\end{itemize}
	Therefore, we get
	\begin{multline*}
		\popprob(u(\fihc_\pi(x))=k_i|\pi) =\sum_{j=l_{i-1}+1}^{l_i-1}\popprob(u(x)=j|\pi) + \\
		+ \popprob(u(x)=l_{i-1},\pi x_{1}=0|\pi)+\popprob(u(x)=l_i,\pi x_{1}=1|\pi)\\
		=\sum_{j=l_{i-1}+1}^{l_i-1}\popprob(u(x)=j) + \popprob(u(x)=l_{i-1},\, x_{1}=0)+\popprob(u(x)=l_i,\, x_{1}=1)\\
		=\sum_{j=l_{i-1}+1}^{l_i-1}\binom{k}{j}2^{-k} + \binom{k-1}{l_{i-1}}2^{-k}+\binom{k-1}{l_i-1}2^{-k}\\
		=\left[\sum_{j=l_{i-1}+1}^{l_i-1}\left(\binom{k-1}{j-1}+\binom{k-1}{j}\right) + \binom{k-1}{l_{i-1}}+\binom{k-1}{l_i-1}\right]2^{-k}\\
		=\left[\sum_{j=l_{i-1}}^{l_i-1}\binom{k-1}{j}\right]2\cdot2^{-k}=\left[\sum_{j=l_{i-1}}^{l_i-1}\binom{k-1}{j}\right]2^{-k+1}.
	\end{multline*}
	Therefore 
	\begin{multline*} 
		\popprob(u(\fihc_\pi(x))=k_i) = \sum_{\pi}\left[\sum_{j=l_{i-1}}^{l_i-1}\binom{k-1}{j}\right]2^{-k+1} \cdot \frac1{k!} \\
		%   =\left[\sum_{j=l_{i-1}}^{l_i-1}\binom{k-1}{j}\right]2^{-k+1}\cdot\sum_{\pi}P(\pi)
		=\left[\sum_{j=l_{i-1}}^{l_i-1}\binom{k-1}{j}\right]2^{-k+1},
	\end{multline*}
	which together with the previous calculations gives us the final formula for~$q_2$.
	
	For computing $q_1$ let us notice that apart from $u(\fihc_\pi(x))=k_i$ we must include $u(\fihc_\pi(x))=0$ (the block of zeros), which guarantees that $\fihc_\pi(x)_{vw}=00$. Thus 
	\begin{multline*}
		q_1=\popprob(\fihc_\pi(x)_{vw}=00)\\
		=\sum_{i=1}^{N} \popprob\big(\fihc_\pi(x)_{vw}=00|u(\fihc_\pi(x))=k_i\big)\, \popprob\big(u(\fihc_\pi(x))=k_i\big)+\\
		+ \popprob(u(\fihc_\pi(x))=0).
	\end{multline*}
	The event $u(\fihc_\pi(x))=0$ has a nonzero probability if and only if $l_0>0$ (i.e., if we have a maximum at unitation $0$). Then 
	\begin{multline*}
		\popprob(u(\fihc_\pi(x))=0) = \sum_{j=0}^{l_0-1} \popprob(u(x)=j)
		+\sum_{\pi} \frac1{k!} \popprob(u(x)=l_0, \pi x_{1}=1|\pi)\\
		=2^{-k}\cdot\sum_{j=0}^{l_0-1}\binom{k}{j}+2^{-k}\cdot\sum_{\pi} \frac1{k!}\binom{k-1}{l_0-1}\\
		=2^{-k}\cdot\left(\sum_{j=0}^{l_0-1}\binom{k}{j} + \binom{k-1}{l_0-1}\right).
	\end{multline*} Using the same argumentation as for $q_2$ we get also \[
	\popprob(\fihc_\pi(x)_{vw}=00|u(\fihc_\pi(x))=k_i)=\frac{\binom{k-2}{k_i}}{\binom{k}{k_i}}=\frac{(k-k_i)(k-k_i-1)}{k(k-1)}
	\]
	for $k_i\neq k-1$. It is possible that $k_N=k-1$, however in this situation we have $\popprob(\fihc_\pi(x)_{vw}=00|u(\fihc_\pi(x))=k-1)=0$ (because we need at least two genes to be zero), so we can use the formula for this case, too. Computing $\popprob(u(\fihc_\pi(x))=k_i)$ goes in the same way as previously.   
\end{proof}
%For each problem length $k$ there are only finitely many sequences $MAX_g$ and $MIN_g$, which completely determine the values of $q_1$ and $q_2$. Hence, as announced, there are only countably many points realisable as distributions of dependent pairs.

As an example we consider the following function designed for blocks of length 12:
\[
g(u) =\begin{cases} 1 & \text{if } u \text{ is odd,}\\
	2 & \text{if } u \text{ is divisible by $4$ and } u<12,\\
	3 & \text{if } u=12,\\
	0 & \text{ otherwise}.
\end{cases}
\]
It will also be studied experimentally in Section \ref{sec:exp} under the name \emph{ridge function}. This function's monotonicity is represented by $MAX_g=(4,8)$ and $MIN_g=(2,6,10).$ Moreover, we have $k_0=0$ and $k_3=12$ (here $N=2$). Using formulas from the theorem, we get 
\begin{eqnarray*}
	q_1 &=& 2^{-11}\cdot \left(\frac{(12-4)(11-4)}{12\cdot11}\sum_{j=2}^{5} \binom{11}{j} + \frac{(12-8)(11-8)}{12\cdot11}\sum_{j=6}^{9} \binom{11}{j} \right)\\ 
	&+& 2^{-12}\cdot\left(\binom{12}{0} + \binom{12}{1} + \binom{11}{1}\right) = \frac{25}{96} = 0.26041(6),\\
	q_2 &=& 2^{-11}\cdot \left(\frac{4\cdot(12-4)}{12\cdot11}\sum_{j=2}^{5} \binom{11}{j} + \frac{8\cdot(12-8)}{12\cdot11}\sum_{j=6}^{9} \binom{11}{j} \right) \\
	&=& \frac{23}{96} = 0.23958(3).
\end{eqnarray*}
As we can see, $q_1 + q_2 = \frac12$, which is not surprising, since the monotonicity of $g$ is symmetric (that is, there is a symmetric function that has the same monotonicity as $g$).

\subsection{Undecidable cases}
For the rest of this section, we will be mostly interested in investigating when the DSM fails to provide perfect decomposition, regardless of the size of the population. 
As was already stated, if genes $x_i$, $x_m$ lie in different blocks, then they are stochastically independent. Recall that two random variables $X$ and $Y$ are stochastically independent if and only if $D(X,Y)=1$. By the Law of Large Numbers and the continuity of $D$, the entries of the DSM will converge to their theoretical values
as we enlarge the population of individuals. So the DSM will be faulty if some pair $x_i$, $x_j$ of dependent genes is stochastically independent. The DSM will not contain \emph{any} information about the dependencies of genes if and only if \emph{all} pairs of genes are stochastically independent. At the end of the section, we will give examples of models illustrating this pathology. Note that this kind of behaviour makes useless all statistical methods, which, on some level of recognizing the dependencies, use the information about the distributions of the pairs of the genes. However, it does not disqualify other statistical methods because pairwise stochastic independence of variables $X_1,X_2,\ldots,X_n$ does not imply independence of larger sets of these variables (in case of concatenations, 
investigating the whole block of dependent variables would reveal that it cannot be stochastically independent, because some combinations of genes are not possible).

Obvioulsy, for any two binary variables $X,Y$ we have $\popprob(X=0)=\popprob(X=0,Y=0)+\popprob(X=0,Y=1)$ and $\popprob(X=1)=\popprob(X=1,Y=0)+\popprob(X=1,Y=1)$. Assume that the joint distribution of $X$ and $Y$ has the form $(q_1,q_2,q_2,q_3)$ (i.e., it lies in the triangle $\mathcal T$). Then we have $\popprob(X=0)=\popprob(Y=0)=q_1+q_2$ and $\popprob(X=1)=\popprob(Y=1)=q_2+q_3$. From that we get that the variables $X$ and $Y$ are independent if and only if:
\begin{equation}
	\label{pato_line_initial_formulas}
	q_1=(q_1+q_2)^2 \quad \textrm{and}\quad
	q_2=(q_1+q_2)(q_2+q_3)\quad \textrm{and}\quad
	q_3=(q_2+q_3)^2.
\end{equation}

\begin{lem}
	\label{pato_line_one_formula}
	If any equality in \eqref{pato_line_initial_formulas} holds, then all three hold.
\end{lem}

\begin{proof}
	Assume that $q_1=(q_1+q_2)^2$. Then $(q_1+q_2)(q_2+q_3)=(q_1+q_2)(q_2+1-2q_2-q_1)=(q_1+q_2)(1-(q_1+q_2))=q_1+q_2-(q_1+q_2)^2=q_1+q_2-q_1=q_2$. Automatically, the third equality also holds. 
	
	Now assume that $q_2=(q_1+q_2)(q_2+q_3)$. Then $(q_2+q_3)^2=(q_2+q_3)(1-q_1-q_2)=q_2+q_3-(q_1+q_2)(q_2+q_3)=q_3$ (to obtain the first equality exchange $q_1$ with $q_3$).
	
	The last case is identical as the first one.
	%Finally, take $q_3=(q_2+q_3)^2$. Then $(q_1+q_2)^2=(1-q_2-q_3)^2=1-2(q_2+q_3)+(q_2+q_3)^2=1-2q_2-q_3=q_1$.
\end{proof}

\begin{lem}
	The set of all points in $\mathcal T$ satisfying the equality $q_1=(q_1+q_2)^2$ (and so the set of all distributions in $\mathcal T$ of two stochastically independent binary variables) is a curve $\mathcal C$, passing through points $(1,0,0,0)$, $(0,0,0,1)$ and $(\frac14,\frac14,\frac14,\frac14)$ (see Figure \ref{fig:curva}). 
\end{lem}
\begin{proof}
	Define $\varphi:\mathcal \R^4\to\mathbb R^2$ by $\varphi(q_1,q_2,q_2,q_3)=(q_1+q_2,\sqrt{2}q_2)$. This is a linear map on $\mathbb R^4$, contracting distances between points in $\mathcal T$ by factor $\frac{1}{\sqrt{2}}$, because
	\begin{multline*}
		\Vert (p_1,p_2,p_2,1-p_1-2p_2)-(q_1,q_2,q_2,1-q_1-2q_2)\Vert^2\\=(p_1-q_1)^2+2(p_2-q_2)^2+(p_1-q_1+2p_2-2q_2)^2\\=2(p_1-q_1)^2+6(p_2-q_2)^2+4(p_1-q_1)(p_2-q_2)\\=2((p_1-q_1+p_2-q_2)^2+2(p_2-q_2)^2)\\=2\Vert(p_1+p_2,\sqrt2p_2)-(q_1+q_2,\sqrt2 q_2)\Vert^2.
	\end{multline*}
	Hence, $\varphi$ just scales the triangle $\mathcal T$ and embeds it into $\mathbb R^2$---its image is the triangle $\overline{\mathcal T}\subset \mathbb R^2$ with vertices $(0,0),(1,0),(\frac12,\frac{\sqrt2} 2)$. 
	
	Consider the parametrization of the set $\mathcal C=\{(q_1,q_2,q_2,q_3)\in \mathcal T: (q_1+q_2)^2=q_1\}$. The condition $(q_1+q_2)^2=q_1$ is equivalent to $q_2=q_1+q_2-(q_1+q_2)^2$, which for points $(x,y)\in \overline{\mathcal{T}}$ translates to $y=\sqrt 2(x-x^2)$. Hence, $\varphi(\mathcal C)=\{(x,y)\in \overline{\mathcal T}: y=\sqrt 2(x-x^2) \}$. Thus we obtain that $\mathcal C$ is a scaled graph of the function on $[0,1]$ given by the formula $\sqrt 2(x-x^2)$. Obviously, the center of the simplex $\mathcal S$ is the apex of the parabola $\mathcal C$.
\end{proof}

Moreover, the center of the simplex is the only common point of $\mathcal C$ and the interval $(0,\frac12,\frac12,0),(\frac12,0,0,\frac12)$ (the height of $\mathcal T$), hence it is the only `fully symmetrical' point, which is also a distribution of independent variables.

\subsection{Examples of undecidability}

From the preceding section, we know that a problem is undecidable by the analysis of the DSM if and only if the distribution of the pair of dependent genes lies on the curve $\mathcal C$. Below, we illustrate it with some examples of the previously postulated form. Interestingly, up to now we have found essentially only one example within the assumed class of problems, whose distribution lies on $\mathcal{C}$ and is different from $(\frac14,\frac14,\frac14,\frac14)$. This gives a potential area for further investigation.

\begin{figure}[ht]
	\centering
	\includegraphics[width=\linewidth] {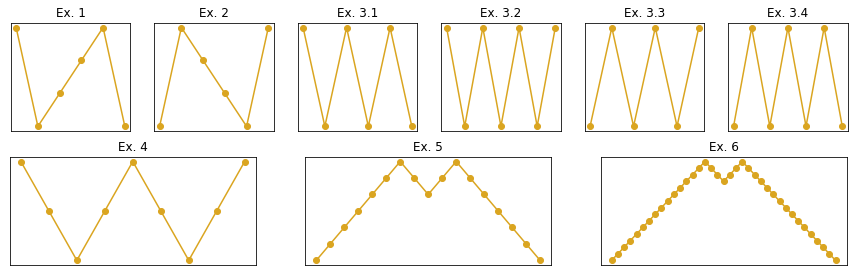}
	\caption{The plots for Examples 1-6.} 
	\label{fig:pato1}
\end{figure}

The distributions of the dependent genes are computed using the formulas from the Theorem \ref{distr_general}. The plots of all the examples are shown in Figure \ref{fig:pato1}.
\begin{itemize}
	\item[1.] We take $k=5$, $MAX_g=(4)$, $MIN_g=(1,5)$. Since $l_0>0$, we have $k_0=0$. Then $(q_1,q_2,q_2,q_3)=(\frac1{16},\frac3{16},\frac3{16},\frac9{16})$ and we easily see that $(q_1+q_2)^2=q_1$.
	\item[2.] The second example is just a mirror of the previous one: we take $k=5$, $MAX_g=(1)$, $MIN_g=(0,4)$, $k_2=5$, and we have $(q_1,q_2,q_2,q_3)=(\frac9{16},\frac3{16},\frac3{16},\frac1{16})$. Taking the mirror of the function (i.e., drawing it backward) just switches the roles of $0$ and $1$, so in the distribution we just switch $q_1$ with $q_3$.
	\item[3.] One can prove the following:  for any $k\geqslant 3$, if we take $MAX_g$, $MIN_g$ such that $l_0\in\{0,1\}$ and $l_{i-1}+1=k_i=l_i-1$ for any $i\geqslant 1$, then $(q_1,q_2,q_2,q_3)=(\frac14,\frac14,\frac14,\frac14)$. If $l_0=0$ then $k_i=2i-1$, $l_i=2i$ and $N=\left\lfloor\frac k2\right\rfloor$. On the other hand, if $l_0=1$ then $k_i=2i$, $l_i=2i+1$ and $N=\left\lceil\frac k2\right\rceil-1$. In both cases one needs to transform the formulas from the Theorem \ref{distr_general}. {It is crucial to notice that now} we have $\sum_{j=l_{i-1}}^{l_i-1} \binom{k-1}{j}=\binom{k-1}{l_{i-1}}+\binom{k-1}{l_{i-1}+1}=\binom{k}{l_{i-1}+1}$. Then it is possible to transform the formula to the form $2^{-k+1}\cdot \sum_{i=0}^{k-3}\binom{k-3}{i}=2^{-k+1}\cdot 2^{k-3}=\frac14$.
	
	The plots of four possible cases of this example, for {$k=5$ or $k=6$}, have been placed on Figure \ref{fig:pato1}. Notice, that for $k=4$ and $l_0=1$ we get a special case of the two types of problems from the Section \ref{sec:bimod_and_revert}---bimodal and reverted bimodal deceptive functions.
\end{itemize}

All the other examples have the distribution $(\frac14,\frac14,\frac14,\frac14)$.
\begin{itemize}
	\item[4.] $k=8$, $MAX_g=(4)$, $MIN_g=(2,6)$, $k_0=0$, $k_2=8$.
	\item[5.] $k=16$, $MAX_g=(6,10)$, $MIN_g=(0,8,16)$.
	\item[6.] $k=36$, $MAX_g=(15,21)$, $MIN_g=(0,18,36)$.
\end{itemize}

We checked by the full search that these are all the {undecidable} examples for $k\in\{3,\ldots,40\}$ and $N\in\{1,2,3,4\}$ (number of maxima on $\{1,\ldots,k-1\}$).

%===================================

\section{Experiments}
\label{sec:exp}

The results of the experiments presented in this section have two objectives. First, we experimentally verify the precision of the proposed estimate. Second, we wish to check if the proposed estimate can be useful in explaining the results of the state-of-the-art SLL-using optimizers.\par

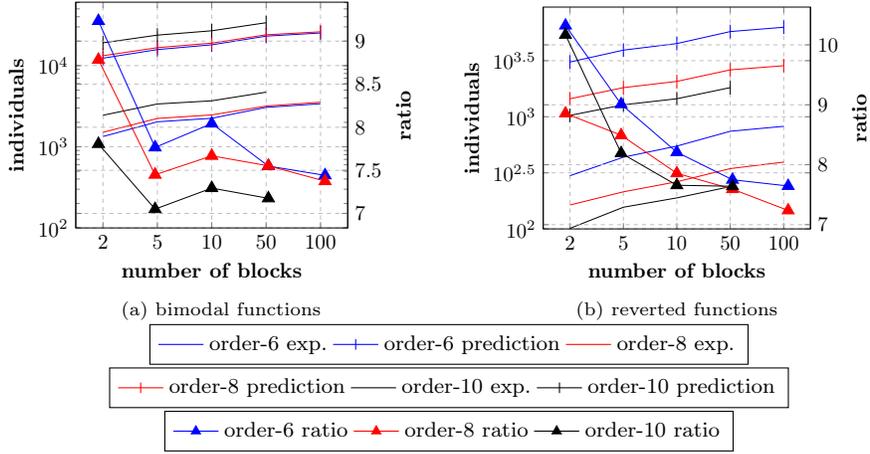
\begin{figure}[ht]
	\centering
	\begin{subfigure}[b]{0.49\linewidth}
		\resizebox{0.95\linewidth}{!}{%
			\tikzset{every mark/.append style={scale=2.5}}
			\begin{tikzpicture}
				\begin{axis}[%
					legend columns=-1,
					legend to name=named,
					legend entries={order-6 exp., order-6 prediction, order-8 exp.},
					%legend to name=named,
					%legend columns=-1,
					%legend entries={LT-GOMEA-DLED,LT-GOMEA-SLL,P3-DLED,P3-SLL},
					%legend to name=named,
					%legend pos=outer north east
					%legend style={at={(0.5,-0.17)},anchor=north,legend cell align=left}
					xtick={1,2,3,4,5},
					xticklabels={2,5,10,50,100},
					xmin=0.5,
					xmax=5.5,
					ymode=log,
					ymin=1e2,
					%ymax=1e9,
					%legend pos=south east,
					xlabel=\textbf{number of blocks},
					ylabel=\textbf{individuals},
					grid,
					grid style=dashed,
					ticklabel style={scale=1.5},
					label style={scale=1.5},
					legend style={font=\fontsize{8}{0}\selectfont}
					]

					\addplot[
					color=blue,
					mark=,
					]
					coordinates {
						(1,1337.6)(2,2030)(3,2243.6)(4,3058.9)(5,3390.2)
					};

					\addplot[
					color=blue,
					mark=|,
					]
					coordinates {
						(1,12351)(2,15765)(3,18046)(4,23092)(5,25229)
						
					};

					\addplot[
					color=red,
					mark=,
					]
					coordinates {
						(1,1507.2)(2,2235)(3,2468.4)(4,3174.4)(5,3540.5)
					};  
					
					\addplot[
					color=red,
					mark=|,
					]
					coordinates {(1,13239)(2,16648)(3,18928)(4,23973)(5,26110)
					};  
					
					\addplot[
					color=black,
					mark=,
					]
					coordinates {
						(1,2444.8)(2,3369)(3,3684.6)(4,4709)
					};   
					
					\addplot[
					color=black,
					mark=|,
					]
					coordinates {
						(1,19084)(2,23750)(3,26874)(4,33786)
					};

				\end{axis}

				\begin{axis}[%
					%axis y line*=right,
					ylabel near ticks, 
					yticklabel pos=right,
					axis x line=none,
					legend columns=-1,
					%legend to name=named,
					%legend columns=-1,
					%legend entries={dgGA,cGOMEA,P3,P3-DLED,LT-GOMEA-DLED},
					legend to name=named,
					legend entries={order-6 ratio, order-8 ratio, order-10 ratio},
					legend to name=namedRatio,
					%xtick={1,2,3,4,5,6},
					%xticklabels={0,1,2,3,4,5},
					%xmin=0.5,
					%xmax=6.5,
					%ymode=log,
					%ymin=1e4,
					%ymax=1e9,
					%legend pos=south east,
					%xlabel=\textbf{unitation},
					ylabel=\textbf{ratio},
					grid,
					grid style=dashed,
					ticklabel style={scale=1.5},
					label style={scale=1.5},
					legend style={font=\fontsize{8}{0}\selectfont}
					]
					
					\addplot[
					color=blue,
					mark=triangle*,
					]
					coordinates {
						(1,9.23370215311005)(2,7.76600985221675)(3,8.04332323052237)(4,7.54911896433358)(5,7.44174384992036)      };

					\addplot[
					color=red,
					mark=triangle*,
					]
					coordinates {
						(1,8.78383757961783)(2,7.44876957494407)(3,7.66812510128018)(4,7.5519783266129)(5,7.37466459539613)
					};

					\addplot[
					color=black,
					mark=triangle*,
					]
					coordinates {
						(1,7.8059554973822)(2,7.0495696052241)(3,7.29360039081583)(4,7.17477171373965)
					};   
					
				\end{axis}
			\end{tikzpicture}
		}
		\caption{bimodal functions}
		\label{fig:modelVSreality:bimodal}
	\end{subfigure}
	\begin{subfigure}[b]{0.49\linewidth}
		\resizebox{0.95\linewidth}{!}{%
			\tikzset{every mark/.append style={scale=2.5}}
			\begin{tikzpicture}
				\begin{axis}[%
					legend columns=-1,
					legend to name=named2,
					legend entries={order-8 prediction, order-10 exp., order-10 prediction},
					xtick={1,2,3,4,5},
					xticklabels={2,5,10,50,100},
					xmin=0.5,
					xmax=5.5,
					ymode=log,
					ymin=1e2,
					%ymax=1e9,
					%legend pos=south east,
					xlabel=\textbf{number of blocks},
					ylabel=\textbf{individuals},
					grid,
					grid style=dashed,
					ticklabel style={scale=1.5},
					label style={scale=1.5},
					legend style={font=\fontsize{8}{0}\selectfont}
					]
					
					\addplot[
					color=red,
					mark=|,
					]
					coordinates {(1,1447)(2,1819)(3,2068)(4,2620)(5,2853)
					};  
					
					\addplot[
					color=black,
					mark=,
					]
					coordinates {
						(1,101.1)(2,156.1)(3,189.1)(4,238.2)
						
					};   
					
					\addplot[
					color=black,
					mark=|,
					]
					coordinates {
						(1,1028)(2,1279)(3,1448)(4,1820)
					};

					\addplot[
					color=blue,
					mark=,
					]
					coordinates {
						(1,298.1)(2,436)(3,547.7)(4,742.4)(5,822.1)
					};

					\addplot[
					color=blue,
					mark=|,
					]
					coordinates {
						(1,3077)(2,3928)(3,4496)(4,5753)(5,6285)
					};

					\addplot[
					color=red,
					mark=,
					]
					coordinates {(1,163.4)(2,214.2)(3,263.2)(4,345.1)(5,394.2)
					};

				\end{axis}

				\begin{axis}[%
					%axis y line*=right,
					ylabel near ticks, 
					yticklabel pos=right,
					axis x line=none,
					legend columns=-1,
					%legend entries={Minimal pop. size until finding optimal TBB set},
					%legend to name=named,
					legend columns=-1,
					%legend entries={LT-GOMEA-DLED,LT-GOMEA-SLL,P3-DLED,P3-SLL},
					%legend to name=named,
					%xtick={1,2,3,4,5,6},
					%xticklabels={0,1,2,3,4,5},
					%xmin=0.5,
					%xmax=6.5,
					%ymode=log,
					%ymin=1e4,
					%ymax=1e9,
					%legend pos=south east,
					%xlabel=\textbf{unitation},
					ylabel=\textbf{ratio},
					grid,
					grid style=dashed,
					ticklabel style={scale=1.5},
					label style={scale=1.5},
					legend style={font=\fontsize{8}{0}\selectfont}
					]
					
					\addplot[
					color=blue,
					mark=triangle*,
					]
					coordinates {
						(1,10.3220395840322)(2,9.0091743119266)(3,8.20887347087822)(4,7.74919181034483)(5,7.64505534606495)   };      
					
					\addplot[
					color=red,
					mark=triangle*,
					]
					coordinates {
						(1,8.85556915544676)(2,8.49206349206349)(3,7.85714285714286)(4,7.59200231816865)(5,7.23744292237443)
						
					};

					\addplot[
					color=black,
					mark=triangle*,
					]
					coordinates {
						(1,10.1681503461919)(2,8.19346572709801)(3,7.65732416710735)(4,7.64063811922754)
					};   
					
				\end{axis}
			\end{tikzpicture}
		}
		\caption{reverted functions}
		\label{fig:modelVSreality:reverted}
	\end{subfigure}
	\ref{named}
	\ref{named2}
	\ref{namedRatio}

	\hspace{0.0001 cm}
	%\ref{named}
	%\ref{named2}

	\caption{Concatenations of bimodal and reverted bimodal functions of different order. The comparison between the 90th percentile of minimal population size necessary to obtain a perfect DSM and the value obtained from the estimation.}
	\label{fig:modelVSreality}
\end{figure}

In Fig. \ref{fig:modelVSreality} we compare the minimal population size of FIHC-optimized individuals necessary for obtaining a perfect DSM returned by the proposed estimation and the experiments. The complete results for all functions considered can be found in Tables \ref{tab:exp_results1} and \ref{tab:exp_results2}. 
A single experiment consisted of creating successive populations by adding new individuals to the population from the previous step until the perfect decomposition was obtained. The size of the final population was taken as the outcome of the experiment. 
Each such experiment was repeated 100 times.
%and was terminated when the generated population of FIHC-optimized individuals was the basis for obtaining a perfect DSM. 
The experiments were terminated if the population size exceeded $5\cdot10^7$ individuals or the experiment execution time exceeded 24 hours on a PowerEdge Dell Server with two AMD EPYC 9654 processors and 1.5TB RAM. In Fig.~\ref{fig:modelVSreality}, we present the comparison between the 90th percentile of the experimental results with predictions given by Theorem \ref{thm:main} for bimodal and reverted bimodal function concatenations. For both functions, we consider orders 6, 8, and 10. Note that our prediction does not directly refer to the method used in experiments (it rather says that if 100 populations of size $s_{min}$ were used to create DSMs than we expect at least 90\% of them to give perfect decomposition, not that the 90th percentile of the obtained growing populations will lie below $s_{min}$).\par

As presented in Fig. \ref{fig:modelVSreality:bimodal}, the populations necessary for obtaining a perfect DSM are almost of the same size for the bimodal-6 and bimodal-8 functions (in between $10^3$ and $10^4$ individuals in the experiments performed). The proposed estimation reports population sizes that are approximately 7 times larger. However, the relations between the estimated and the experimental values seem the same: the population sizes are very close for bimodal-6 and bimodal-8, while the population size for bimodal-10 is larger. Note that except for problems of very small length (consisting of only two blocks), the ratio remains on the same level between 7 and 8.\par

\begin{table}[ht] 
	\caption{Comparison of the estimated sizes of populations with the values from the experiments for bimodal deceptive function. Note that the problem length is equal to the number of blocks times the order of the problem.}
	\scriptsize
	\begin{tabular}{l|rrrr} \label{tab:exp_results1}
		& problem   & 90\% percentile        & 0.9 probability      &   \\
		& length   & (experiments)       & (estimation)     & ratio  \\
		\hline
		bimodal-6      & 12  & 1337.60    & 12351   & 9.23  \\
		& 30  & 2030.00    & 15765   & 7.77  \\
		& 60  & 2243.60    & 18046   & 8.04  \\
		& 300 & 3058.90    & 23092   & 7.55  \\
		& 600 & 3390.20    & 25229   & 7.44  \\
		bimodal-8      & 16  & 1507.20    & 13239   & 8.78  \\
		& 40  & 2235.00    & 16648   & 7.45  \\
		& 80  & 2468.40    & 18928   & 7.67  \\
		& 400 & 3174.40    & 23973   & 7.55  \\
		& 800 & 3540.50    & 26110   & 7.37  \\
		bimodal-10     & 20  & 2444.80    & 19084   & 7.81  \\
		& 50  & 3369.00    & 23750   & 7.05  \\
		& 100 & 3684.60    & 26874   & 7.29  \\
		bimodal-12     & 24  & 3684.60    & 28027   & 7.61  \\
		& 60  & 4670.30    & 34611   & 7.41  \\
		& 120 & 5466.90    & 39020   & 7.14  \\
		bimodal-16     & 32  & 7424.80    & 54071   & 7.28  \\
		& 80  & 9499.40    & 66033   & 6.95  \\
		& 160 & 10833.60   & 74049   & 6.84  \\
		bimodal-20     & 40  & 12465.10   & 90403   & 7.25  \\
		& 100 & 16688.00   & 109539  & 6.56  \\
		& 200 & 19073.40   & 122368  & 6.42  \\
		bimodal-24     & 48  & 21216.40   & 137008  & 6.46  \\
		& 120 & 25849.80   & 165024  & 6.38  \\
		bimodal-30     & 60  & 35171.10   & 226672  & 6.44  \\
		& 150 & 43763.70   & 271175  & 6.20  \\
		bimodal-50     & 100 & 116403.70  & 705101  & 6.06  \\
		& 500 & 279276.40  & 917096  & 3.28  \\
		bimodal-100    & 200 & 2201928.00 & 3200016 & 1.45  
	\end{tabular}
\end{table}

\begin{table}[htbp] 
	\caption{Comparison of the estimated sizes of populations with the values from the experiments for reverted bimodal deceptive function}
	\scriptsize
	\begin{tabular}{l|rrrr} \label{tab:exp_results2}
		& problem   & 90\% percentile        & 0.9 probability      &   \\
		& length   & (experiments)       & (estimation)     & ratio  \\
		\hline
		rev. bimodal-6   & 12  & 298.10     & 3077    & 10.32 \\
		& 30  & 436.00     & 3928    & 9.01  \\
		& 60  & 547.70     & 4496    & 8.21  \\
		& 300 & 742.40     & 5753    & 7.75  \\
		& 600 & 822.10     & 6285    & 7.65 \\
		rev. bimodal-8   & 16  & 163.40     & 1447    & 8.86  \\
		& 40  & 214.20     & 1819    & 8.49  \\
		& 80  & 263.20     & 2068    & 7.86  \\
		& 400 & 345.10     & 2620    & 7.59  \\
		& 800 & 394.20     & 2853    & 7.24  \\
		rev. bimodal-10  & 20  & 101.10     & 1028    & 10.17 \\
		& 50  & 156.10     & 1279    & 8.19  \\
		& 100 & 189.10     & 1448    & 7.66  \\
		& 500 & 238.20     & 1820    & 7.64  \\
		rev. bimodal-12  & 24  & 91.10      & 841     & 9.23  \\
		& 60  & 129.10     & 1038    & 8.04  \\
		& 120 & 150.20     & 1170    & 7.79  \\
		& 600 & 204.00     & 1463    & 7.17  \\
		rev. bimodal-16  & 32  & 73.10      & 667     & 9.12  \\
		& 80  & 97.00      & 814     & 8.39  \\
		& 160 & 118.20     & 913     & 7.72  \\
		& 800 & 156.00     & 1132    & 7.26  \\
		rev. bimodal-20  & 40  & 67.00      & 584     & 8.72  \\
		& 100 & 89.10      & 708     & 7.95  \\
		& 200 & 104.00     & 790     & 7.60  \\
		rev. bimodal-30  & 60  & 53.10      & 490     & 9.23  \\
		& 150 & 71.10      & 587     & 8.26  \\
		& 300 & 82.00      & 651     & 7.94  \\
		rev. bimodal-24  & 48  & 58.10      & 535     & 9.21  \\
		& 120 & 86.10      & 645     & 7.49  \\
		& 240 & 87.20      & 718     & 8.23  \\
		rev. bimodal-50  & 100 & 37.10      & 425     & 11.46 \\
		& 250 & 63.00      & 501     & 7.95  \\
		& 500 & 73.00      & 552     & 7.56  \\
		rev. bimodal-100 & 200 & 16.10      & 378     & 23.48 \\
		rev. bimodal-200 & 400 & 18.10      & 354     & 19.56
	\end{tabular}
\end{table}

The results for the reverted function concatenations are different than for bimodal functions. However, the observation that the estimated values are highly similar to the experimental ones but approximately 7-9 times higher remains valid. Thus, the results of the performed verification indicate that the proposed estimation is relatively reliable in showing the differences in the minimal population size necessary for obtaining a perfect DSM for various problems. The above analysis suggests that it may be reasonable to consider  values obtained from the proposed estimations as an ingredient of a measure of difficulty in decomposing a given problem with SLL.\par

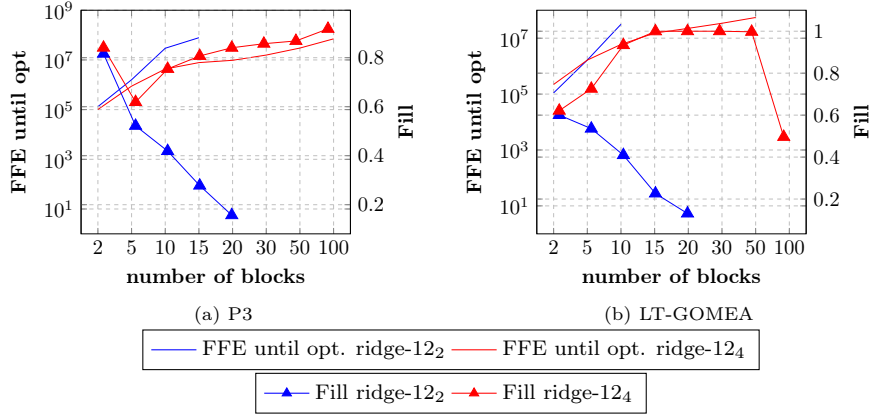
\begin{figure}[ht]
	\centering
	\begin{subfigure}[b]{0.49\linewidth}
		\resizebox{0.95\linewidth}{!}{%
			\tikzset{every mark/.append style={scale=2.5}}
			\begin{tikzpicture}
				\begin{axis}[%
					legend columns=-1,
					legend to name=named3,
					legend entries={FFE until opt. ridge-$12_2$, FFE until opt. ridge-$12_4$},
					%legend to name=named,
					%legend columns=-1,
					%legend entries={LT-GOMEA-DLED,LT-GOMEA-SLL,P3-DLED,P3-SLL},
					%legend to name=named,
					%legend pos=outer north east
					%legend style={at={(0.5,-0.17)},anchor=north,legend cell align=left}
					xtick={1,2,3,4,5,6,7,8},
					xticklabels={2, 5, 10, 15, 20, 30, 50, 100},
					xmin=0.5,
					xmax=8.5,
					ymode=log,
					ymin=1,
					ymax=1e9,
					%legend pos=south east,
					xlabel=\textbf{number of blocks},
					ylabel=\textbf{FFE until opt},
					grid,
					grid style=dashed,
					ticklabel style={scale=1.5},
					label style={scale=1.5},
					legend style={font=\fontsize{8}{0}\selectfont}
					]

					\addplot[
					color=blue,
					mark=,
					]
					coordinates {
						(1,131652)(2,1594251)(3,29546956)(4,78415210)
					};

					\addplot[
					color=red,
					mark=,
					]
					coordinates {
						(1,101064.5)(2,880720)(3,4304621.5)(4,7800621)(5,9660763)(6,15710928)(7,30056045)(8,70536823)
						
					};

				\end{axis}

				\begin{axis}[%
					%axis y line*=right,
					ylabel near ticks, 
					yticklabel pos=right,
					axis x line=none,
					legend columns=-1,
					%legend to name=named,
					%legend columns=-1,
					%legend entries={dgGA,cGOMEA,P3,P3-DLED,LT-GOMEA-DLED},
					%legend to name=named,
					%legend columns=-1,
					legend entries={$\Fill$ ridge-$12_2$, $\Fill$ ridge-$12_4$},
					legend to name=namedFILL,
					%xtick={1,2,3,4,5,6},
					%xticklabels={0,1,2,3,4,5},
					%xmin=0.5,
					%xmax=6.5,
					%ymode=log,
					%ymin=1e4,
					%ymax=1e9,
					%legend pos=south east,
					%xlabel=\textbf{unitation},
					ylabel=\textbf{Fill},
					grid,
					grid style=dashed,
					ticklabel style={scale=1.5},
					label style={scale=1.5},
					legend style={font=\fontsize{8}{0}\selectfont}
					]
					
					\addplot[
					color=blue,
					mark=triangle*,
					]
					coordinates {
						(1,0.81628788)(2,0.5219697)(3,0.418939395)(4,0.278030305)(5,0.15719697)
					};
					
					\addplot[
					color=red,
					mark=triangle*,
					]
					coordinates {
						(1,0.84090909)(2,0.618939395)(3,0.754166665)(4,0.80707071)(5,0.840530305)(6,0.857070705)(7,0.868409095)(8,0.917689395)
						
					};

				\end{axis}

			\end{tikzpicture}
		}
		\caption{P3}
		\label{fig:optimizers:p3}
	\end{subfigure}
	\begin{subfigure}[b]{0.49\linewidth}
		\resizebox{0.95\linewidth}{!}{%
			\tikzset{every mark/.append style={scale=2.5}}
			\begin{tikzpicture}
				\begin{axis}[%
					%legend entries={Rev-6 exp., Rev-6 model, Rev-8 exp.,Rev-8 model, Rev-10 exp., Rev-10 model},
					%legend to name=named,
					legend columns=-1,
					%legend entries={LT-GOMEA-DLED,LT-GOMEA-SLL,P3-DLED,P3-SLL},
					%legend to name=named,
					%legend pos=outer north east
					%legend style={at={(0.5,-0.17)},anchor=north,legend cell align=left}
					xtick={1,2,3,4,5,6,7,8},
					xticklabels={2, 5, 10, 15, 20, 30, 50, 100},
					xmin=0.5,
					xmax=8.5,
					ymode=log,
					ymin=1,
					ymax=1e8,
					%legend pos=south east,
					xlabel=\textbf{number of blocks},
					ylabel=\textbf{FFE until opt},
					grid,
					grid style=dashed,
					ticklabel style={scale=1.5},
					label style={scale=1.5},
					legend style={font=\fontsize{8}{0}\selectfont}
					]

					\addplot[
					color=blue,
					mark=,
					]
					coordinates {
						(1,110433.5)(2,1648853.5)(3,31367874.5)
					};

					\addplot[
					color=red,
					mark=,
					]
					coordinates {
						(1,221564.5)(2,1644223.5)(3,5963806.5)(4,15214804.5)(5,22511652.5)(6,34178118.5)(7,54977658)
					};

				\end{axis}

				\begin{axis}[%
					%axis y line*=right,
					ylabel near ticks, 
					yticklabel pos=right,
					axis x line=none,
					legend columns=-1,
					%legend entries={Minimal pop. size until finding optimal TBB set},
					%legend to name=named,
					legend columns=-1,
					%legend entries={LT-GOMEA-DLED,LT-GOMEA-SLL,P3-DLED,P3-SLL},
					%legend to name=named,
					%xtick={1,2,3,4,5,6},
					%xticklabels={0,1,2,3,4,5},
					%xmin=0.5,
					%xmax=6.5,
					%ymode=log,
					%ymin=1e4,
					ymax=1.1,
					%legend pos=south east,
					%xlabel=\textbf{unitation},
					ylabel=\textbf{Fill},
					grid,
					grid style=dashed,
					ticklabel style={scale=1.5},
					label style={scale=1.5},
					legend style={font=\fontsize{8}{0}\selectfont}
					]
					
					\addplot[
					color=blue,
					mark=triangle*,
					]
					coordinates {
						(1,0.60037879)(2,0.53560606)(3,0.409469695)(4,0.22651515)(5,0.131060605)
					};      
					
					\addplot[
					color=red,
					mark=triangle*,
					]
					coordinates {
						(1,0.62121212)(2,0.725757575)(3,0.93371212)(4,0.99949495)(5,1)(6,1)(7,0.99719697)(8,0.49625)
						
					};

				\end{axis}
			\end{tikzpicture}
		}
		\caption{LT-GOMEA}
		\label{fig:optimizers:ltGom}
	\end{subfigure}
	\ref{named3}
	\ref{namedFILL}
	%\ref{namedPerc}

	\hspace{0.0001 cm}
	%\ref{named}
	%\ref{named2}

	\caption{SLL-using optimizers (P3 and LT-GOMEA) in solving SLL-hard (ridge-$12_2$) and not SLL-hard (ridge-$12_4$) function concatenations. The dependency between FFE necessary for finding an optimal solution and linkage quality.}.
	\label{fig:optimizers}
\end{figure}

In this work we identify certain functions whose concatenations can not be successfully decomposed by the populations of FIHC-optimized individuals, i.e., even arbitrarily large populations will not yield a perfect DSM. 
We will call them SLL-undecidable. 

By a \emph{noised} version of a problem $f$ we understand a problem of the form $\tilde f(x)=f(x)+n(x)$, such that:
\begin{itemize}
	\item $\tilde f$ has the same globally optimal solutions as $f$,
	\item $\tilde f$ has more locally optimal solutions than $f$.
\end{itemize}
In \cite{3lo} a noised bimodal function $b(u(x)) + n(u(x))$ was considered with $u(x)$ being unitation, $b(u)$ is the bimodal function of order 10 and
\[
n(u) = \begin{cases}
	-2 & \textrm{for } u=5,\\
	-1 & \textrm{for } u=0,3,7,10,\\
	\phantom{-} 0 & \textrm{for } u= 1,4,6,9,\\
	\phantom{-} 1 & \textrm{for } u= 2,8.
\end{cases}
\]
It turned out that for this function it is much harder to find optimum than for an analogous bimodal function. From our analysis of examples it follows that this function is in fact SLL-undecidable---it falls into the case~3 of our examples in the preceding section.
Therefore, we wish to check if there are other problems that have similar features.\par

For even $k$ we define {ridge-}$k_2:\{0,1,\ldots,k\}\to\{0,1,2\}$ as follows:
$$\text{ridge-}k_2(u)=\begin{cases} 0 & \text{if } u \text{ is odd,}\\
	1 & \text{if } u \text{ is even and } u<k,\\
	2 & \text{if } u=k.
\end{cases}
$$
and {ridge-}$k_4:\{0,1,\ldots,k\}\to\{0,1,2,3\}$ as follows:
$$\text{ridge-}k_4(u)=\begin{cases} 1 & \text{if } u \text{ is odd,}\\
	2 & \text{if } u \text{ is divisible by $4$ and } u<k,\\
	3 & \text{if } u=k,\\
	0 & \text{ otherwise}.
\end{cases}
$$
In Fig. \ref{fig:optimizers}, we show the results of optimization of ridge-$12_2$ and ridge-$12_4$. Both functions are similar and have the same size. The difference is that ridge-$12_2$ (in fact, ridge-$k_2$ for any $k$, as it falls into the case 3 of our examples) is SLL-undecidable, while ridge-$12_4$ is not, because $q_1\neq \frac 14 = (q_1 + q_2)^2$, as we calculated in the preceding section (compare formula \eqref{pato_line_initial_formulas}). (Note that ridge-$8_4$ is again SLL-undecidable, as it is in the class of our example~4 in the preceding section.) We consider two SLL-using optimizers, P3 and LT-GOMEA. The computation budget was $10^8$ fitness function evaluations (FFE), and each experiment was repeated 30 times. We report the median FFE necessary for finding the optimal solution. Additionally, we report the value of the highest $\Fill$ measure value that refers to one of the DSMs maintained by the optimizer at the end of the run, i.e., we compute the $\Fill$ measure for each DSM maintained for each pyramid level in P3 and for each DSM maintained for each subpopulation in LT-GOMEA. Then, we choose the highest $\Fill$ values and report it (see Section \ref{sec:rw} for the definition of $\Fill$). \par

As presented in Fig.~\ref{fig:optimizers:p3}, P3 solves much larger instances of ridge-$12_4$ than in the case of ridge-$12_2$. Additionally, linkage quality at the end run is always higher (frequently significantly higher) for ridge-$12_4$ concatenations despite a higher budget spent on optimizing ridge-$12_2$ concatenations. Note that the linkage quality drops down even for the 240-bit test cases that have used the whole FFE budget without finding the optimal solution. The results of LT-GOMEA reported in Fig. \ref{fig:optimizers:ltGom} lead to the same conclusions. Thus, we can state that the performed analysis explains why P3 and LT-GOMEA are significantly more effective in solving ridge-$12_4$ concatenations than in the case of ridge-$12_2$.

\section{Conclusions}
\label{sec:conc}

In this work, we propose the estimation of the minimal population size necessary to find a perfect DSM for the concatenations of symmetric functions of unitation. The experimental results confirm that the proposed approach may suffice to indicate which problems are difficult to solve for state-of-the-art SLL-using optimizers. The proposed analysis leads to finding other functions that are hard to decompose by SLL and improves the understanding of the results obtained using these optimizers. It also allows us to identify the weaknesses of the SLL-using optimizers. Thus, it allows them to improve by mitigating their weaknesses. The main future work directions will be the further improvement of the proposed estimation precision and taking into account the overlapping and non-symmetric problems.

\biboptions{sort}
\bibliographystyle{ieeetr}
\bibliography{sizes}

\end{document}